\documentclass{article}

\usepackage{arxiv}

\usepackage[utf8]{inputenc} 
\usepackage[T1]{fontenc}    
\usepackage{hyperref}       
\usepackage{url}            
\usepackage{booktabs}       
\usepackage{amsfonts}       
\usepackage{nicefrac}       
\usepackage{microtype}      
\usepackage{lipsum}		
\usepackage{graphicx}
\usepackage{natbib}
\usepackage{doi}
\usepackage{amsmath,amssymb}
\DeclareMathOperator*{\argmax}{argmax} 
\usepackage{mathtools}
\usepackage[ruled,vlined]{algorithm2e}
\SetKw{KwBy}{by}
\SetKwInput{KwInput}{Input}
\SetKwInput{KwOutput}{Output}
\usepackage[flushleft]{threeparttable}
\usepackage{longtable}
\usepackage{diagbox}
\usepackage{chngpage}

\usepackage{pdflscape}
\usepackage{float}
\usepackage{multirow}
\usepackage{subcaption}
\usepackage{hhline}
\usepackage[english]{babel}
\newcommand{\defeq}{\vcentcolon=}
\newtheorem{theorem}{Theorem}

\title{Penalized  Deep Partially Linear Cox Models with Application to CT Scans of Lung Cancer Patients}

\author{Yuming Sun\\
	Department of Mathematics\\
	William \& Mary, Williamsburg\\
	\texttt{ysun30@wm.edu} \\
	\And
	Jian Kang\\
    Department of Biostatistics\\
	University of Michigan, Ann Arbor\\
	\texttt{jiankang@umich.edu}\\
	\And
        Chinmay Haridas\\
        Division of Thoracic Surgery, Department of Surgery\\
        Massachusetts General Hospital\\
        \texttt{chinmayharidas25@gmail.com}\\
        \And
        Nicholas R. Mayne\\
        Department of Medicine\\
        Duke University\\
        \texttt{chinmayharidas25@gmail.com}\\
        \And
        Alexandra L. Potter\\
        Division of Thoracic Surgery, Department of Surgery\\
        Massachusetts General Hospital\\
        \And
        Chi-Fu Jeffrey Yang\\
        Division of Thoracic Surgery, Department of Surgery\\
        Massachusetts General Hospital\\
        \texttt{cjyang@mgh.harvard.edu}\\
        \And
        David C. Christiani\\
        Department of Environmental Health and Epidemiology\\
        Harvard T.H. Chan School of Public Health\\
        \texttt{dchris@hsph.harvard.edu}
        \And
	Yi Li\\
	Department of Biostatistics\\
	University of Michigan, Ann Arbor\\
	\texttt{yili@umich.edu}
}

\date{}

\hypersetup{
pdftitle={A template for the arxiv style},
pdfsubject={q-bio.NC, q-bio.QM},
pdfauthor={David S.~Hippocampus, Elias D.~Striatum},
pdfkeywords={First keyword, Second keyword, More},
}

\begin{document}
\maketitle

\begin{abstract}
Lung cancer is a leading cause of cancer mortality globally, highlighting the importance of understanding its mortality risks to design effective patient-centered therapies. The National Lung Screening Trial (NLST) employed computed tomography texture analysis, which provides objective measurements of texture patterns on CT scans, to quantify the mortality risks of lung cancer patients. Partially linear Cox models have gained popularity for survival analysis by dissecting the hazard function into parametric and nonparametric components, allowing for the effective incorporation of both well-established risk factors (such as age and clinical variables) and emerging risk factors (e.g., image features) within a unified framework. However,  when the dimension of parametric components exceeds the sample size, the task of model fitting becomes formidable, while nonparametric modeling grapples with the curse of dimensionality. We propose a novel Penalized Deep Partially Linear Cox Model (Penalized DPLC), which incorporates the SCAD penalty to select important texture features and employs a deep neural network to estimate the nonparametric component of the model. We prove the convergence and asymptotic properties of the estimator and compare it to other methods through extensive simulation studies, evaluating its performance in risk prediction and feature selection. The proposed method is applied to the NLST study dataset to uncover the effects of key clinical and imaging risk factors on patients' survival. Our findings provide valuable insights into the relationship between these factors and survival outcomes.
\end{abstract}


\section{Introduction}
\label{sec:intro}
Even with the advent of modern medicine, lung cancer mortality remains high, with a 5-year survival rate lower than 20\% among advanced patients \citep{bade2020lung}.  Identifying risk factors relevant to lung cancer survival is essential for designing cancer prevention programs \citep{barbeau2006results} for prevention and early detection.  
The National Lung Cancer Screen Trial (NLST) was designed to investigate the use of  computed tomography (CT) for lung cancer detection, enrolling more than 53,000 participants  from August 2002 through April 2004 with about 26,000  randomly assigned to receive  CT~\citep{national2011reduced}. In addition, clinical information, such as age, gender, smoking history, and cancer stage, was collected for each patient. The study found a 20\% decrease in lung cancer mortality for patients screened by CT.
It is of interest to examine whether  CT  confers valuable features to help predict lung cancer survival and design efficient disease management strategies.
CT texture analysis provides objective assessments of the texture patterns of the tumor by evaluating the 
relationship of voxel intensities~\citep{lubner2017ct}.
Identifying reproducible and robust texture features in the presence of other clinical factors affecting patients' outcomes remains a challenge due to the sensitivity of radiomic features to factors such as scanner type, segmentation, and organ motion~\citep{lambin2017radiomics}. 

Partially linear Cox models have gained popularity as a useful extension of the classic Cox models \citep{Cox1972regression} for survival analysis. This model offers more flexibility in the risk function by separating the hazard function into parametric relative risks for certain covariates and nonparametric relative risks for the others \citep{huang1999efficient}. In the NLST analysis, we have chosen to adopt this model by assigning the parametric risks to the texture features and the nonparametric risks to the clinical features such as age, gender, and race. This setup provides a clear interpretation of texture features as in regular Cox models, facilitates the selection of crucial radiomic features, and offers extra flexibility in modeling the effects and potential interactions of the well-known clinical features.

To estimate the nonparametric risk function, researchers have proposed various methods, such as 
polynomial splines \citep{huang1999efficient}. Recently, \citet{zhong2022deep} made a breakthrough by using deep neural networks (DNN) to estimate the nonparametric risk function in partially linear Cox models and established an optimal minimax rate of convergence for the DNN-based estimator, and showed that DNN  approximates a wide range of nonparametric functions with faster convergence. However, the performance of this method remains unknown when dealing with a large number of texture features, which is the case in the NLST study.
 
In many applications, the neural network has proven to be  powerful for approximating complex functions by providing accurate approximations of continuous functions~\citep{ leshno1993multilayer}.  Under some smoothness and structural assumptions, \citet{schmidt2020nonparametric} showed that DNN estimators may circumvent the
curse of dimensionality and achieve the optimal minimax rate of convergence.
With limited samples, however, a complex DNN can still lead to overfitting~\citep{li2020gradient,srivastava2014dropout}. Various methods, such as early stopping during training~\citep{li2020gradient}, and adding dropout layers~\citep{srivastava2014dropout}, have been proposed to address overfitting, 
but these methods have not been widely studied in the survival context.
 
To fill this gap, we propose a Penalized Deep Partially Linear Cox Model (Penalized DPLC). This framework identifies valuable radiomic features and models the complex relationships between survival outcomes and established clinical characteristics such as age, body mass index (BMI), and pack-years of smoking. The main contributions of our work lie in the proposed penalized estimation, in the context of DNN,   to select texture features that influence survival outcomes while avoiding overfitting, combining feature selection and deep learning in one solution. Second, we demonstrate the asymptotic properties of the estimator by determining its convergence rates and proving selection consistency.
Finally, we perform comprehensive simulations to validate the proposed model's theoretical properties and compare it with the other methods in risk prediction and feature selection.

The paper is structured as follows. Section 2 introduces the Penalized DPLC model and the penalized log partial likelihood and  Section 3 presents an efficient alternating optimization algorithm. Theoretical results are provided in Section 4, where we prove the convergence rate and variable selection consistency. In Section 5, we conduct simulations to evaluate the performance of the Penalized DPLC and compare it with other state-of-the-art models. We apply the Penalized DPLC to a dataset from the NLST study in Section 6 to identify important texture features related to patient survival and find that the selected features are clinically interpretable and align with the previous  findings.

\section{SCAD-penalized Deep Partially Linear Cox Models}
A partially linear Cox model assumes a hazard function:
\begin{equation}\label{deep partially linear Cox model}
    \lambda(t|\mathbf{x},\mathbf{z}) = \lambda_0(t)\exp(\boldsymbol{\beta}_0^{\top}\mathbf{x} + g_0(\mathbf{z})),
\end{equation}
where $\mathbf{x} \in \mathbb{R}^p$ and $\mathbf{z} \in  \mathbb{R}^r$ are
two covariate vectors, and  $\lambda_0(t)$ is the baseline hazard. This class of models contains the Cox proportional hazards model as a special case if $g_0(\mathbf{z})$ is a linear function of $\mathbf{z}$.
 In NLST, $\mathbf{x}$ represents texture features and $\mathbf{z}$ represents known clinical features such as age,  BMI, gender,  race and cancer stage. The coefficients measuring the impact of texture features are represented by $\boldsymbol{\beta}_0$, while the non-parametric risk function of clinical features is represented by $g_0$ and is to be approximated by a function in a deep neural network (DNN). We consider a practical setting where $p$, the dimension of $\mathbf{x}$, can exceed the sample size, which necessitates variable selection. As such, $\boldsymbol{\beta}_0$ is an $s_{\beta}$-sparse vector, i.e., $\Vert\boldsymbol{\beta}_0\Vert_0 = s_{\beta} < p$. On the other hand, the important clinical features have a moderate dimension of $r$, and their complex impacts   are to be modeled by a DNN. 
  
 

 As defined in ~\citet{schmidt2020nonparametric} and \citet{zhong2022deep}, a DNN with architecture $(L,\mathbf{p})$ has $L + 1$ layers, including an input layer, $L-1$ hidden layers and an output layer, and a width vector $\mathbf{p} = (p_1,p_2,\dots,p_{L+1})$ whose elements are the numbers of neurons in the corresponding layer. In this context, a DNN has  two or more hidden layers, while  shallow networks are those with only one hidden layer ~\citep{schmidt2020nonparametric}. In our case, the dimension of the input features, $p_1=r,$ and the dimension of output, $p_{L+1}=1$. An $(L+1)$-layered neural network with an architecture $(L,\mathbf{p})$ can be expressed as a composite function, $g:$ $\mathbb{R}^{r} \to \mathbb{R}^{1}$,  with $L$ folds, i.e., 
$
    g=g_{L} \circ g_{L-1} \circ \cdots \circ g_{1},
$
where `$\circ$' is the functional composition, and the $l$th fold function, 
$g_l(\cdot) = \sigma_{l}(\mathbf{W}_{l}\cdot +\mathbf{b}_{l}): \mathbb{R}^{p_{l}} \to \mathbb{R}^{p_{l+1}}  \, \, {\rm with} \, \, l=  1,\ldots,L.
$
Here,  $\mathbf{W}_{l}$ is a ${p_{l+1}\times p_{l}}$ weight matrix, $\mathbf{b}_{l}$ is a $p_{l+1}$-dimensional bias vector and `$\cdot$'  represents an input from layer $l$. We use $\Theta$ to denote the set of parameters for the neural network containing all the weight matrices and bias vectors to be estimated. The function $\sigma_{l}: \mathbb{R}^{p_{l+1}}\to \mathbb{R}^{p_{l+1}}$ is an activation function, possibly nonlinear, that operates component-wise on a vector.

 Various activation functions exist, with ReLU, i.e., $\mbox{max}(0,\mathbf{a})$,  being a commonly used choice.  Our primary emphasis lies in neural networks employing ReLU functions across all layers, although these can be readily modified.
  Moreover, DNNs with complex network structures and a large number of parameters are prone to overfitting. 
 This work concentrates on a class of DNNs with sparsity constraints on the weight and bias matrices~\citep{zhong2022deep,schmidt2020nonparametric}:
  $$
    \mathcal{G}(L,\mathbf{p},s,G) = \{g \in \mathcal{G}(L,\mathbf{p}): \sum_{l=1}^{L}\Vert\mathbf{W}_l\Vert_0 + \Vert\mathbf{b}_l\Vert_0 \leq s, \Vert g\Vert_{\infty} \leq G \}.
    $$
Here,  $s \in \mathbb{N}_+$ (the set of positive integers),  $G > 0$, $\Vert g \Vert_{\infty}= \sup\{|g(z)|: z \in \mathbb{D} \subset \mathbb{R}^r\}$ is the sup-norm of  function $g$, and $\mathbb{D}$ is a bounded set. 
In implementation, directly specifying or determining $s$, which controls network sparsity, is not the norm. Instead, a commonly employed technique is  a ``dropout" procedure within the hidden layers, which  randomly removes hidden neurons with a defined probability, referred to as the dropout rate~\citep{srivastava2014dropout}.  To determine an appropriate dropout rate, we conduct a grid search as done in our later simulations and data analysis.


With right censoring, we let $U_i$ and $C_i$ denote the survival and censored times for subject $i$, respectively. We observe $ T_i = \min(U_i, C_i)$, and $\Delta_i = 1(U_i \leq C_i)$, where $1(\cdot)$ is the indicator function, 
and assume the observed data $\mathcal{D} = \{(T_i,\Delta_i,\mathbf{x}_i,\mathbf{z}_i), i = 1,\dots,n\}$ {are independently and identically distributed (IID).} 
 To estimate $g_0$ in (\ref{deep partially linear Cox model}), we suggest using a DNN, denoted as $\mathcal{G}(L,\mathbf{p},s,\infty)$, which takes $\mathbf{z} \in \mathbb{R}^r$ as input features and produces a scalar output. To achieve variable selection among $\mathbf{x}$, we propose a penalized estimation approach.

To proceed, we define the partial likelihood as
\begin{equation}\label{partial likelihood}
    \ell(\boldsymbol{\beta},g) = \frac{1}{n}\sum_{i=1}^n \Delta_i\Big[\boldsymbol{\beta}^{\top}\mathbf{x}_i + g(\mathbf{z_i})- \log\Big\{ \sum_{j\in R_i}\exp\big(\boldsymbol{\beta}^{\top}\mathbf{x}_j + g(\mathbf{z_j})\big)\Big\}\Big],
\end{equation}
where $R_i =\{j:T_j \ge T_i\}$, the at-risk set at time $T_i$, and $g \in \mathcal{G}(L,\mathbf{p},s,\infty)$. We would estimate $\boldsymbol{\beta}$ and $g(\cdot)$ by maximizing (\ref{partial likelihood}), where, to accommodate sparsity, we  propose  to use the SCAD penalty~\citep{fan2001variable,fan2002variable} defined as
 \[   p'_{\lambda}(|\beta|) =   \lambda \Big\{\mathrm{I}(|\beta| \leq \lambda) + \frac{(a\lambda - |\beta|)_+}{(a - 1)\lambda }\mathrm{I}(|\beta|>\lambda)\Big\}, \qquad a>2,
 \]
yielding a penalized log partial likelihood, $  PL(\boldsymbol{\beta},g) =
 \ell(\boldsymbol{\beta},g) - \sum_{j=1}^p p_{\lambda}(|\beta_j|).$  The SCAD penalty is indeed a quadratic spline function with knots at $\lambda$ and $a\lambda$, where  $\lambda>0$ is viewed
as the tuning parameter controlling the sparsity of $\boldsymbol{\beta}$, and is assumed to converge to 0 as $n \rightarrow \infty$, though for simplification we omit its dependence on $n$.   

We estimate $(\boldsymbol{\beta}_0,g_0)$  by maximizing $PL(\boldsymbol{\beta},g)$, or,  equivalently, minimizing the loss function
which is  defined as the negative penalized log partial likelihood:
  \begin{equation}\label{loss function}
    Q(\boldsymbol{\beta},g)= q(\boldsymbol{\beta},g) + \sum_{j=1}^p p_{\lambda}(|\beta_j|),
\end{equation}
where $q(\boldsymbol{\beta},g)=- \ell(\boldsymbol{\beta},g)$. That is, the estimate of $(\boldsymbol{\beta}_0,g_0)$ is obtained via 
\begin{equation}{\label{loss function2}}
    (\widehat{\boldsymbol{\beta}},\widehat{g}) = \arg\min_{\boldsymbol{\beta},g\in \mathbb{R}^p \times \mathcal{G}} Q(\boldsymbol{\beta},g).
\end{equation}

We present below an optimization algorithm for solving (\ref{loss function2}) alternately, which  uses the adaptive moment estimation (Adam) algorithm to estimate $g$ given an estimate of $\boldsymbol{\beta}$, and, subsequently,  uses the resulting estimate $\widehat{g}$ to estimate $\boldsymbol{\beta}$ via coordinate descent. 

\begin{enumerate}
    \item[Step 1.] Initialize $\boldsymbol{\beta}$ with $ \boldsymbol{\widehat{\beta}}^{(0)}$.
    \item[Step 2.] Denote by  $\boldsymbol{\widehat{\beta}}^{(k-1)}$  the estimate of $\boldsymbol{\beta}$ at the $(k-1)$th iteration. Solve (\ref{loss function2}) for $g$, with
     $\boldsymbol{\beta}$ fixed at $\boldsymbol{\widehat{\beta}}^{(k-1)}$,
     by using Adam (Algorithm \ref{adam}), where   $\widehat{g}^{(k)}$ denotes the current estimate.
    \item[Step 3.] With $g$ fixed at $\widehat{g}^{(k)}$, solve (\ref{loss function2}) for $\boldsymbol{\beta}$ by using the coordinate descent algorithm (Algorithm \ref{coordinate descent}), where  $\boldsymbol{\widehat{\beta}}^{(k)}$ denotes the estimate at the current iteration.
\end{enumerate}

We repeat Steps 2 and 3 until convergence. In Step 2, we employ an adapted Adam algorithm (Algorithm 1), a form of stochastic gradient descent \citep{kingma2014adam}, to estimate $\Theta$  (the weight matrices and bias vectors) in the neural network. The algorithm is adaptive as the update of $\Theta$ at each iteration step 
stems from adaptive estimation of the first and second moments of the stochastic gradients of the empirical loss  \citep{kingma2014adam}. We initialize the biases to be 0 and use   \textit{Xavier initialization} to  initialize the weights ~\citep{glorot2010understanding}. 
To ensure numerical stability, we add a small $\epsilon_0>0$  to the denominator, and the update for each parameter is determined by the adaptive estimates for the first and second moments of the gradients of the empirical loss at each iteration.
Algorithm 1 distinguishes from the traditional Adam method in that it updates the parameters in the neural network while fixing $\boldsymbol{\beta}$ at its previous iteration, rather than updating all parameters simultaneously. When implementing Algorithm \ref{adam}, we do not require convergence with a given update of $\boldsymbol{\beta}$. In our experience, several iterative steps would be sufficient. 
Also as a large number of iterations may lead to overfitting of DNN, early stopping may prevent overfitting and can produce a  consistent network~\citep{ji2021early}.

\begin{algorithm}
\SetKwInOut{Output}{Output}
\SetKwInOut{Input}{Input}
\caption{Adam in alternating optimization}\label{adam}
\Input{$r_1$, $r_2$, $\gamma$, $\widehat{\boldsymbol{\beta}}^{(k-1)}$, $\iota$ 
}

Initialize $m^{(0)} \gets 0$, $v^{(0)} \gets 0$, $t \gets 1$, $\Theta^{(0)}$\\
\While{$\Vert \widehat{\Theta}^{(t)} -  \widehat{\Theta}^{(t-1)} \Vert_2 >\iota$}{
    $m^{(t)} \gets r_1 \cdot m^{(t-1)} + (1 - r_1) \cdot \nabla_{\Theta}Q(\widehat{\boldsymbol{\beta}}^{(k-1)},\widehat{g}^{(t)})$
    \\
    $v^{(t)} \gets r_2 \cdot m^{(t-1)} + (1 - r_2) \cdot \{\nabla_{\Theta}Q(\widehat{\boldsymbol{\beta}}^{(k-1)},\widehat{g}^{(t)})\}^2$\\
    $\widehat{m}^{(t)} \gets m^{(t)}/(1 - r_1^t),$ $\widehat{v}^{(t)} \gets v^{(t)}/(1 - r_2^t)$\\
    $\widehat{\Theta}^{(t)} \gets \widehat{\Theta}^{(t-1)} - \gamma \widehat{m}^{(t)}/(\sqrt{\widehat{v}^{(t)}} + \epsilon_0)$ \\
    $t \gets t + 1$
    }
\Output{
$\widehat{g}^{(k)} \gets g(\cdot \mid \widehat\Theta^{(t)})$
}
{\rm Note: the square, division and square root from lines 3 to 6 are operated elementwise.}
\end{algorithm}

Step 3 carries out  a coordinate descent algorithm. The advantage of coordinate descent is that the parameters, $\boldsymbol{\beta}$, are updated individually, where the closed-form solution for each parameter is available, greatly facilitating  the computation~\citep{breheny2011coordinate}. Specifically, let $\boldsymbol{\xi} = \mathbf{X}\boldsymbol{\beta} \in \mathbb{R}^{n}$, where $\mathbf{X} =(\mathbf{x}_1, \ldots, \mathbf{x}_n)^\top$ is the covariate ($\mathbf{x}$) matrix of the
$n$ subjects in the data. We denote the gradient and Hessian of the function $q$ with respect to $\boldsymbol{\beta}$ and $\boldsymbol{\xi}$  
given the current estimate of the neural network, $\widehat{g}^{(k)}$, as $q'(\boldsymbol{\beta};\widehat{g}^{(k)})$, $q''(\boldsymbol{\beta};\widehat{g}^{(k)})$, $q'(\boldsymbol{\xi};\widehat{g}^{(k)})$, and $q''(\boldsymbol{\xi};\widehat{g}^{(k)})$. To simplify notation, we will omit $\widehat{g}^{(k)}$ in the following. The function $q(\boldsymbol{\beta})$ is approximated using a second order Taylor expansion around $\widehat{\mathbf{b}}^{(t)}$:
\begin{align*}
    q(\boldsymbol{\beta}) &\approx q(\widehat{\mathbf{b}}^{(t)}) + (\boldsymbol{\beta} - \widehat{\mathbf{b}}^{(t)})^{\top}q'(\widehat{\mathbf{b}}^{(t)}) + (\boldsymbol{\beta} -\widehat{\mathbf{b}}^{(t)})^{\top}q''(\widehat{\mathbf{b}}^{(t)})(\boldsymbol{\beta} - \widehat{\mathbf{b}}^{(t)})/2\\
    &=\frac{1}{2}(y(\widehat{\boldsymbol{\xi}}^{(t)}) - \boldsymbol{\xi})^{\top}q''(\widehat{\boldsymbol{\xi}}^{(t)})(y(\widehat{\boldsymbol{\xi}}^{(t)}) - \boldsymbol{\xi}) + C(\widehat{\boldsymbol{\xi}}^{(t)},\widehat{\mathbf{b}}^{(t)}),
\end{align*}
where $y(\widehat{\boldsymbol{\xi}}^{(t)}) = \widehat{\boldsymbol{\xi}}^{(t)} - q''(\widehat{\boldsymbol{\xi}}^{(t)})^{-1}q'(\widehat{\boldsymbol{\xi}}^{(t)})$ and $C(\widehat{\boldsymbol{\xi}}^{(t)},\widehat{\mathbf{b}}^{(t)})$ does not depend on $\boldsymbol{\beta}$. The equalities hold as $q'({\boldsymbol{\beta}})= \mathbf{X}^\top
q'({\boldsymbol{\xi}})$ and  $q''({\boldsymbol{\beta}})= \mathbf{X}^\top
q''({\boldsymbol{\xi}}) \mathbf{X}$ by the chain rule.
Then the loss function (\ref{loss function}) at iteration $t$ can be approximated by the penalized weighted sum of squares,
    $Q(\boldsymbol{\beta}) \approx \frac{1}{2}(y(\widehat{\boldsymbol{\xi}}^{(t)}) - \boldsymbol{\xi})^{\top}q''(\widehat{\boldsymbol{\xi}}^{(t)})(y(\widehat{\boldsymbol{\xi}}^{(t)}) - \boldsymbol{\xi}) + C(\widehat{\boldsymbol{\xi}}^{(t)},\widehat{\boldsymbol{\beta}}^{(t)}) + \sum_{j=1}^pp_{\lambda}(|\beta_j|).$ To speed up the algorithm, we may replace $q''(\widehat{\boldsymbol{\xi}}^{(t)})$ by a diagonal matrix, $\mathbf{W}(\widehat{\boldsymbol{\xi}}^{(t)})$, with the diagonal entries of $q''(\widehat{\boldsymbol{\xi}}^{(t)})$:
\begin{align*}
    \mathbf{W}(\widehat{\boldsymbol{\xi}}^{(t)})_{m,m} = q''(\widehat{\boldsymbol{\xi}}^{(t)})_{m,m} = \frac{1}{n}\sum_{i \in C_m}\Delta_i\Bigg \{ \frac{e^{\widehat{\xi}^{(t)}_{m} + \widehat{g}^{(k)}_m} \sum_{j \in R_i} e ^{\widehat{\xi}_j^{(t)}+ \widehat{g}^{(k)}_j}-(e^{\widehat{\xi}_m^{(t)}+ \widehat{g}^{(k)}_m})^2}{(\sum_{j \in R_i} e ^{\widehat{\xi}_j^{(t)}+ \widehat{g}^{(k)}_j})^2} \Bigg \},
\end{align*}
where $C_m = \{i : T_i \leq T_m \}$. In this case,
\begin{align*}
    y(\widehat{\boldsymbol{\xi}}^{(t)})_m = \widehat{\boldsymbol{\xi}}_m^{(t)} +\frac{1}{n\mathbf{W}(\widehat{\boldsymbol{\xi}}^{(t)})_{m,m}}\Bigg \{ \Delta_m - \sum_{i\in C_m} \Delta_i\Bigg( \frac{e^{\widehat{\xi}_m^{(t)} + \widehat{g}^{(k)}_m}}{\sum_{j \in R_i}e^{\widehat{\xi}_j^{(t)} + \widehat{g}^{(k)}_j}} \Bigg)\Bigg\}.
\end{align*}

In the iteration of coordinate descent, the parameters are updated individually; each parameter has a closed-form solution, making the computation manageable. We employ an adaptive rescaling technique  \citep{breheny2011coordinate}; 
the following SCAD-thresholding operator  returns the univariate solution for the SCAD-penalized optimization:
 \[   f_{SCAD}(h,v;a,\lambda) = 
    \begin{cases}
        \frac{S(h,\lambda)}{v}, & \text{if }|h| \leq 2\lambda\\
        \frac{S(h,a\lambda/(a-1))}{v(1 - 1/(a -1))},  & \text{if } 2\lambda < |h| \leq a\lambda\\
        h/v, & \text{if } |h| > a\lambda,
    \end{cases}
\]
where $S(\cdot, \lambda)$ is the soft-thresholding operator with a threshold parameter, $\lambda>0$~\citep{donoho1994ideal}, i.e., $S(h,\lambda) = sign(h)(|h| - \lambda)_+$. Here, the sign function $sign(h)$ equals $ h/|h|$ if $h\ne 0$, and 0 if $h=0$;  $(h)_+= \max(h,0)$. Let $\mathbf{r} = y(\boldsymbol{\xi})-
\boldsymbol{\xi}
$ and $v_j =  \mathbf{x}_j^{\top} \mathbf{W}(\boldsymbol{\xi}) \mathbf{x}_j$. We  define the following input at the $t$th iteration, i.e., 
$
    h_j = \mathbf{x}_j^{\top}\mathbf{W}(\widehat{\boldsymbol{\xi}}^{(t)})\mathbf{r} + v_j \beta_j^{(t)}.
$
The coordinate descent algorithm is presented in Algorithm~\ref{coordinate descent}.

\begin{algorithm}
\SetKwInOut{Output}{Output}
\SetKwInOut{Input}{Input}  
\caption{Coordinate Descent in alternating optimization}\label{coordinate descent}
\Input{$a$, $\lambda$, $\widehat{\mathbf{b}}^{(0)} = \widehat{\boldsymbol{\beta}}^{(k-1)}$, $\widehat{g}^{(k)}$, $\iota$}
Initialize $t \gets 1$, $\widehat{\boldsymbol{\xi}}^{(0)} \gets \mathbf{X}\widehat{\mathbf{b}}^{(0)}$, and $\mathbf{r} \gets y(\widehat{\boldsymbol{\xi}}^{(0)})- \widehat{\boldsymbol{\xi}}^{(0)}$\\
\While{$\Vert \widehat{\mathbf{b}}^{(t)} - \widehat{\mathbf{b}}^{(t-1)}  \Vert_2 > \iota$}{
    \For{$j\gets1$ \KwTo $p$}{
        $h_j \gets \mathbf{x}_j^{\top}\mathbf{W}(\widehat{\boldsymbol{\xi}}^{(t-1)})\mathbf{r} + v_j \beta^{(t-1)}_j$\\
        $\widehat{\beta}^{(t)}_j \gets f_{SCAD}(h_j,v_j; a,\lambda)$\\
        $\mathbf{r} \gets \mathbf{r} - (\widehat{\beta}^{(t)}_j - \widehat{\beta}^{(t - 1)}_j)\mathbf{x}_j$ 
    }
     $\widehat{\boldsymbol{\xi}}^{(t)} \gets  \mathbf{X}\widehat{\mathbf{b}}^{(t)}$\\
     $t \gets t + 1$
    }
\Output{ 
$\widehat{\boldsymbol{\beta}}^{(k)} \gets \widehat{\mathbf{b}}^{(t)}$ }
\end{algorithm}

\section{Regularity Conditions and Statistical Properties}
We impose sparsity on $\boldsymbol{\beta}_0 = (\beta_{10},\dots,\beta_{p0})^{\top} = (\boldsymbol{\beta}_{10}^{\top},\boldsymbol{\beta}_{20}^{\top})^{\top}$ by, without loss of generality,  assuming $\boldsymbol{\beta}_{20} = \mathbf{0}$. We restrict the true nonparametric function $g_0$ to belong to a composite H\"{o}lder class of smooth functions, $\mathcal{H}(q,\alpha,\mathbf{d},\widetilde{\mathbf{d}},M)$, where  the $q$ composition functions are H\"{o}lder smooth functions with parameters $\alpha=(\alpha_1, \ldots,\alpha_q)$ (the orders of smoothness) and $M$ (bound). The concept of the composite H\"{o}lder smooth function has been widely used to facilitate the discussion of the theoretical properties of DNN~\citep{schmidt2020nonparametric,zhong2022deep}. Here, $\mathbf{d} =(d_1, \ldots,d_q)$ and $\widetilde{\mathbf{d}}= (\widetilde{d}_1, \ldots,\widetilde{d}_q)$ are two types of dimension parameters;  the former is the dimension of input at each `layer,' while  the latter  quantifies the \textit{intrinsic dimension} of the arguments of activation functions at each   layer~\citep{zhong2022deep}, often much smaller than the feature dimension at each layer. We will prove that the convergence rate of DNN depends on $\mathbf{\widetilde{d}}$, instead of $\mathbf{d}$, meaning a faster convergence rate than the other nonparametric estimators. Details 
can be found in the Supplement Materials.

Throughout, $\mathbb{E}$ denotes the expectation of random variables; unless otherwise specified,  for any function (random or nonrandom) $f$ and a random vector, $\mathbf{v}$, we define $\mathbb{E}\{f(\mathbf{v})\} 
    \defeq \int f(\mathbf{t}) f_\mathbf{v}(\mathbf{t})d \mathbf{t},$ where $f_\mathbf{v}(\cdot)$    is the density function of $\mathbf{v}$. Thus, the expectation is taken with respect to only the arguments of the $f$ function.
    For a vector $\mathbf{a}$, define $||\mathbf{a}||= (\mathbf{a}^\top \mathbf{a})^{1/2}$, and for a function $g$, define $\Vert g \Vert_{L^2}^2 = \mathbb{E}\{g^2(\mathbf{z})\}$.
    We denote $\widetilde{\alpha}_i = \alpha_i \prod_{k=i+1}^q(\alpha_k \wedge 1 )$ and $\gamma_n = \max_{i = 1, \dots,q} n^{-\widetilde{\alpha}_i/(2\widetilde{\alpha}_i + \widetilde{d}_i)}$, and  assume the following.
\begin{enumerate}
    \item[1.] Considering a class of $s$-sparse DNNs or $\mathcal{G}(L,\mathbf{p},s,G)$, we assume
    $L = O(\log n)$, $s = O(n\gamma_n^2\log n)$ 
    and $n\gamma_n^2  < \min_{l=1,\dots,L} p_l \leq \max_{l=1,\dots,L} p_l < n$.

    \item[2.] {With slightly overuse of notation, denote by $\mathbf{x}$ and $\mathbf{z}$
    the random vectors underlying the observed IID copies of  $\mathbf{x}_i$ and $\mathbf{z}_i$, respectively.}
    Assume $(\mathbf{x}^\top,\mathbf{z}^\top)^\top$ take values in a bounded subset, $\mathbb{D}$, of $\mathbb{R}^{p+r}$ with a joint probability density function bounded away from zero, and $\boldsymbol{\beta}_0$ lies in a compact set, i.e., $\boldsymbol{\beta}_0 \in \{\boldsymbol{\beta} \in \mathbb{R}^p: \Vert\boldsymbol{\beta}\Vert\leq B\}$.
     
    \item[3.] Assume that the nonparametric function $g_0$ belongs to a mean 0 composite H\"{o}lder smooth class, i.e., $g_0 \in \mathcal{H}_0  \coloneqq \{ g \in \mathcal{H}(q,\alpha,\mathbf{d},\widetilde{\mathbf{d}},M) : \mathbb{E}\{g(\mathbf{z})\}= 0\}$  
    and the matrix $\mathbb{E}\{ \mathbf{x} - \mathbb{E}(\mathbf{x} | \mathbf{z})\}^{\otimes 2}$ is nonsingular, where $\mathbf{a}^{\otimes2} = \mathbf{a}\mathbf{a}^{\top}$ for a column vector $\mathbf{a}$. 
    \item[4.] Let $\tau < \infty$ be the maximal followup time.  We assume that   there exits a  $\delta > 0$ such that $P(\Delta =1 |\mathbf{x},\mathbf{z}) > \delta$ and $P(U >  \tau |\mathbf{x},\mathbf{z}) > \delta$ almost surely.
  \end{enumerate}

Condition 1 restricts the architecture of neural networks, balancing the network's flexibility with the estimation accuracy 
\citep{zhong2022deep}. 
Condition 2 is commonly assumed for semiparametric partially linear models~\citep{horowitz2009semiparametric}. The H\"{o}lder smoothness in  Condition 3 
ensures that the function can be approximated by a DNN, while  the zero expectation assumption  yields the identifiability of the deep partially linear Cox model~\citep{zhong2022deep}.  
In Condition 4, $P(\Delta =1 |\mathbf{x},\mathbf{z}) > \delta$ specifies that there is  non-zero probability of observing an event, and $P(U > \tau |\mathbf{x},\mathbf{z}) > \delta$ ensures that there  is  non-zero probability that some subjects are still alive at the end of the study,  both of which guarantee that the partially linear Cox model can be estimated using the observed data.


With
    $a_n = \max\{ p_{\lambda}'(|\beta_{j0}|: \beta_{j0}\neq 0)\}$ and $
    b_n = \max\{ p_{\lambda}''(|\beta_{j0}|: \beta_{j0}\neq 0)\},$
the following theorem establishes the existence and the convergence rates of $\widehat{\boldsymbol{\beta}}$ and $\widehat{g}$.

\begin{theorem}\label{g consistency}
     Under Conditions 1-4, and if $b_n \rightarrow 0$ (with properly chosen $\lambda$), then there exists a local maximizer $(\widehat{\boldsymbol{\beta}},\widehat{g})$ of $PL(\boldsymbol{\beta},g)$ satisfying $\mathbb{E}\{\widehat{g}(\mathbf{z})\} = 0$, such that
    \[
    \Vert \widehat{\boldsymbol{\beta}} - \boldsymbol{\beta}_0 \Vert  = O_p(\gamma_n\log ^2n + a_n), \,\, \, 
    \Vert\widehat{g} - g_0\Vert_{L^2} = O_p(\gamma_n\log ^2n + a_n).
    \]
\end{theorem}

\noindent {\em Remark 1:} The theorem shows that the rate of convergence does not depend on the number of input features, but rather on the intrinsic dimension and smoothness of the function $g_0$, unlike other nonparametric estimators whose convergence rate also depends on the feature dimension. As a result, the DNN estimator may have an advantage when the intrinsic dimension of the true function is low.

We now show that the estimator $\widehat{\boldsymbol{\beta}} = (\widehat{\boldsymbol{\beta}}_{1}^{\top},\widehat{\boldsymbol{\beta}}_{2}^{\top})^{\top}$ 
for  $({\boldsymbol{\beta}}_{10}^{\top},{\boldsymbol{\beta}}_{20}^{\top} =\mathbf{0}^{\top}  )^{\top}$ 
possesses a selection consistency property, that is, $\widehat{\boldsymbol{\beta}}_2 = \mathbf{0}$ with probability going to 1.

\begin{theorem}\label{sparsity}
   Assume that $\lim\inf_{n\rightarrow\infty}\lim\inf_{u\rightarrow0^+}p_{\lambda}'(u)/\lambda>0$, and  $\lambda$ is chosen such that $a_n = O(\gamma_n\log^2n)$, and
        $\lambda \min (n^{1/2}, \{\gamma_n \log^{2}(n)\}^{-1}) \rightarrow \infty$, and the conditions of Theorem \ref{g consistency} hold. Then with probability tending to 1, the estimator $\widehat{\boldsymbol{\beta}}$ in Theorem \ref{g consistency} must satisfy $\widehat{\boldsymbol{\beta}}_2 = \mathbf{0}$.
\end{theorem}

\noindent {\em Remark 2:} Both theorems apply to a broad range of penalty functions. In particular, 
 as shown in 
\cite{fan2001variable}, the SCAD penalty function satisfies 
$\lim\inf_{n\rightarrow\infty}\lim\inf_{u\rightarrow0^+}p_{\lambda}'(u)/\lambda>0$,
and as $\lambda \rightarrow 0+$, $a_n=0$ when $n$ is sufficiently large.
Consequently, if $\lambda$ converges to 0 at an appropriate rate, the SCAD function guarantees  both the convergence rates (Theorem \ref{g consistency}) and variable selection consistency (Theorem \ref{sparsity}).

\section{Simulations}
We conducted simulations to assess the finite sample performance of
our proposed estimator by comparing it with the SCAD-penalized Cox Model~\citep{fan2002variable}, SCAD-penalized Partially Linear Cox Model using polynomial splines~\citep{hu2013variable}, Cox Boosting~\citep{binder2009boosting}, Random Forest~\citep{ishwaran2008random} and Deep Survival Model~\citep{katzman2018deepsurv}. 
 For $i = 1,\dots,n$, we generated  $(\mathbf{x}_i,\mathbf{z}_i)$  from a multivariate Gaussian distribution,  
$
 \mathcal{N}_{p + r}\Bigg \{\mathbf{0},
\big(\begin{smallmatrix}
1 & 0.2 & \dots & 0.2\\
\vdots & \vdots &\ddots & \vdots\\
0.2 & 1 & \dots  & 1
\end{smallmatrix}\big)
\Bigg \},
$
and then generated the true survival time $U_i$ from an exponential distribution with a hazard $  \mu \exp(\boldsymbol{\beta}^T_0 \mathbf{x}_i + g_0(\mathbf{z}_i)),$
where $\mu$ was tuned to adjust for censoring rate and  $\boldsymbol{\beta}_0 \in \mathbb{R}^{p}$ was a sparse vector simulated from the uniform distribution. The number of nonzero elements in $\boldsymbol{\beta}_0$ was $s_{\beta}$, chosen to be much less than the dimension of  $\boldsymbol{\beta}_0$. The censored time $C_i$ was simulated from  $\mathcal{U}[0, {\cal C}]$, where ${\cal C}$ was chosen so that the censoring rate in the simulated data is around 30\%.

We simulated data sets with varying sample sizes and feature sizes. Specifically, we fixed the clinical feature size, $r$, at 8 and the number of nonzero radiomic features, $s_{\beta}$, to be 10, while varying the training sample size, $n$, to be 500 or 1,500 and radiomic feature size, $p$, to be 600 or 1,200. 
We assessed the performance of the model under these four scenarios with different numbers of training samples and feature sizes.  For each simulation setup or configuration, a total of 500 independently simulated datasets were generated. 

We set $g_0:\mathbb{R}^8 \rightarrow \mathbb{R}$ to be a linear or nonlinear function, respectively. That is, 
$g_0(\mathbf{z}) = \boldsymbol{\alpha}_0^{\top}\mathbf{z}$
with  $\boldsymbol{\alpha}_0 \in \mathbb{R}^8$  generated from $\mathcal{U}(-2,2)$
or $ 0.68  \exp(z_1) - 0.45 \log\{ (z_2 - z_3)^2 \} + 0.32  \sin(z_4  z_5) - 0.45  (z_6 - z_7 + z_8)^2 - 0.32$.
We tuned parameters for each method on each  simulated dataset.  Specifically, to identify the neural network structure, we tuned the number of hidden layers and the number of neurons in the hidden layers over a grid of values, i.e., 1 to 4 for the number of hidden layers and  2 to 8 for the number of neurons in the hidden layers, and tuned the dropout rate and the learning rate  from 0.3 to 0.5 and  from 0.005 to 0.02, respectively.
For the SCAD penalty, we set $a = 3.7$ as suggested by \cite{fan2001variable}  and  used grid search over $[0.05,5]$  to find the best $\lambda$  based on the Bayesian Information Criterion (BIC): 
$
-2n\ell(\widehat{\boldsymbol{\beta}},\widehat{g}) + \log n \cdot \widehat{s}_{\beta},$ 
where $\widehat{s}_{\beta}$ is the number of  nonzero coefficient estimates;  
for illustration, Figures S.1(a)-(b) in the Supplement Material  display the selection of $\lambda$  for SCAD-Penalized  DPLC on ten simulated datasets with $(n,p)=(500, 1,200)$ and the solution path for $\widehat{\boldsymbol{\beta}}$ with one randomly selected dataset. 
We tuned for Cox Boosting by determining the penalty value that yielded an optimal count of boosting steps (with a maximum of 200). For Random Forest, we  tuned the terminal node size  from 1 to 150.

\begin{figure}
    \centering
    \includegraphics[scale =0.3]{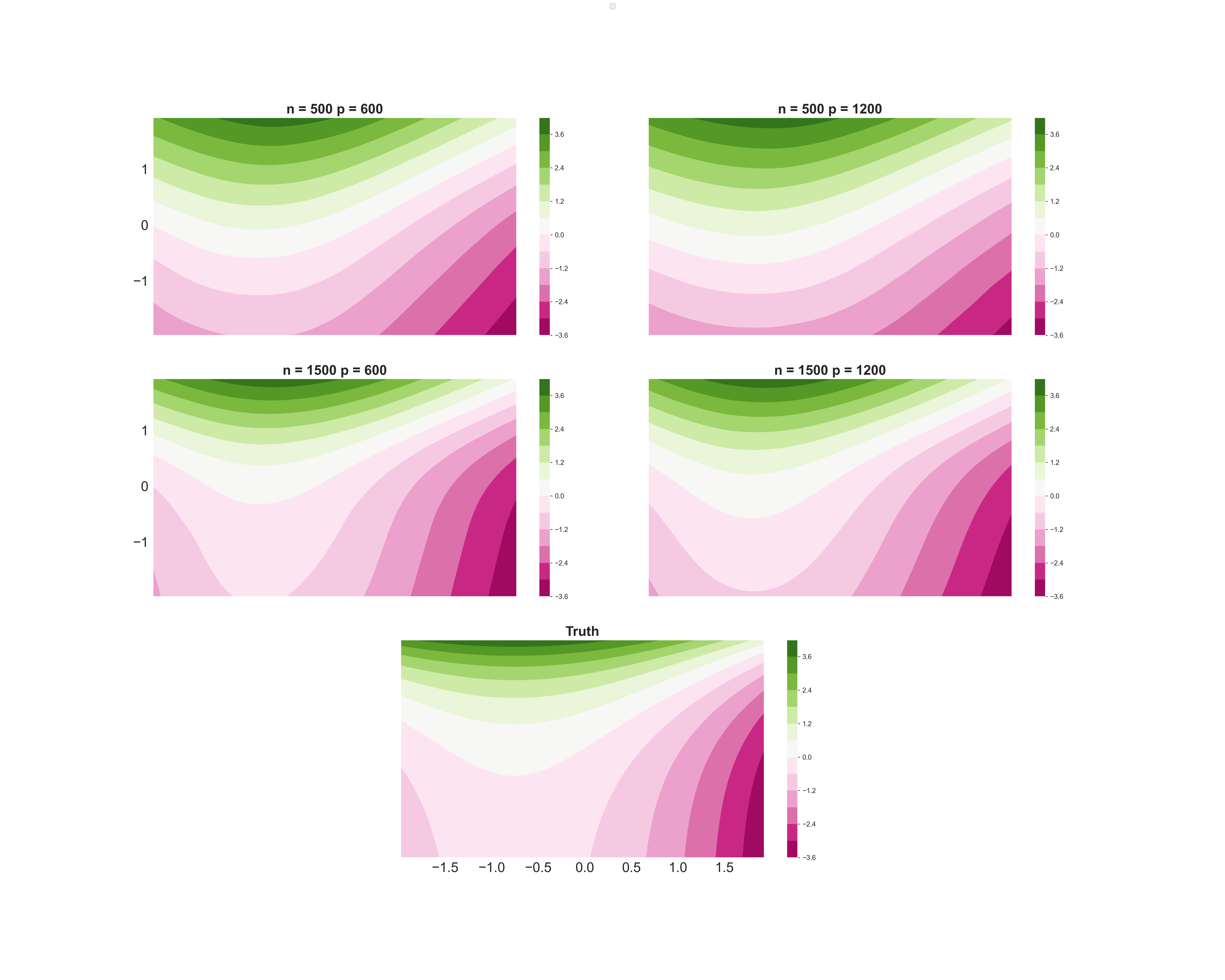}
    \caption{\textbf{The Average Estimates of the Nonlinear Function using 500 Simulated Datasets with Varying $n,p$.}   The plots are made by varying the first two arguments fixing the other six arguments.}
    \label{estimate g}
\end{figure}

 To visually evaluate the accuracy of the DNN estimator in approximating   $g_0$ when it is nonlinear, Figure \ref{estimate g} displays contour plots of the true function and the average DNN estimates based on 500 simulated datasets with $n,p$ varying from 500 to 1,500 and  from 600 to 1,200, respectively. 
 When creating these plots, we fixed the values of the last six arguments of the function at their population means and varied the first two arguments. The results indicate that the DNN estimates provided a good approximation of the true function, with increasing accuracy observed as $n$ increased for a fixed value of $p$.

\begin{figure}
    \centering
    \subfloat[Linear Case]{\includegraphics[scale =0.21]{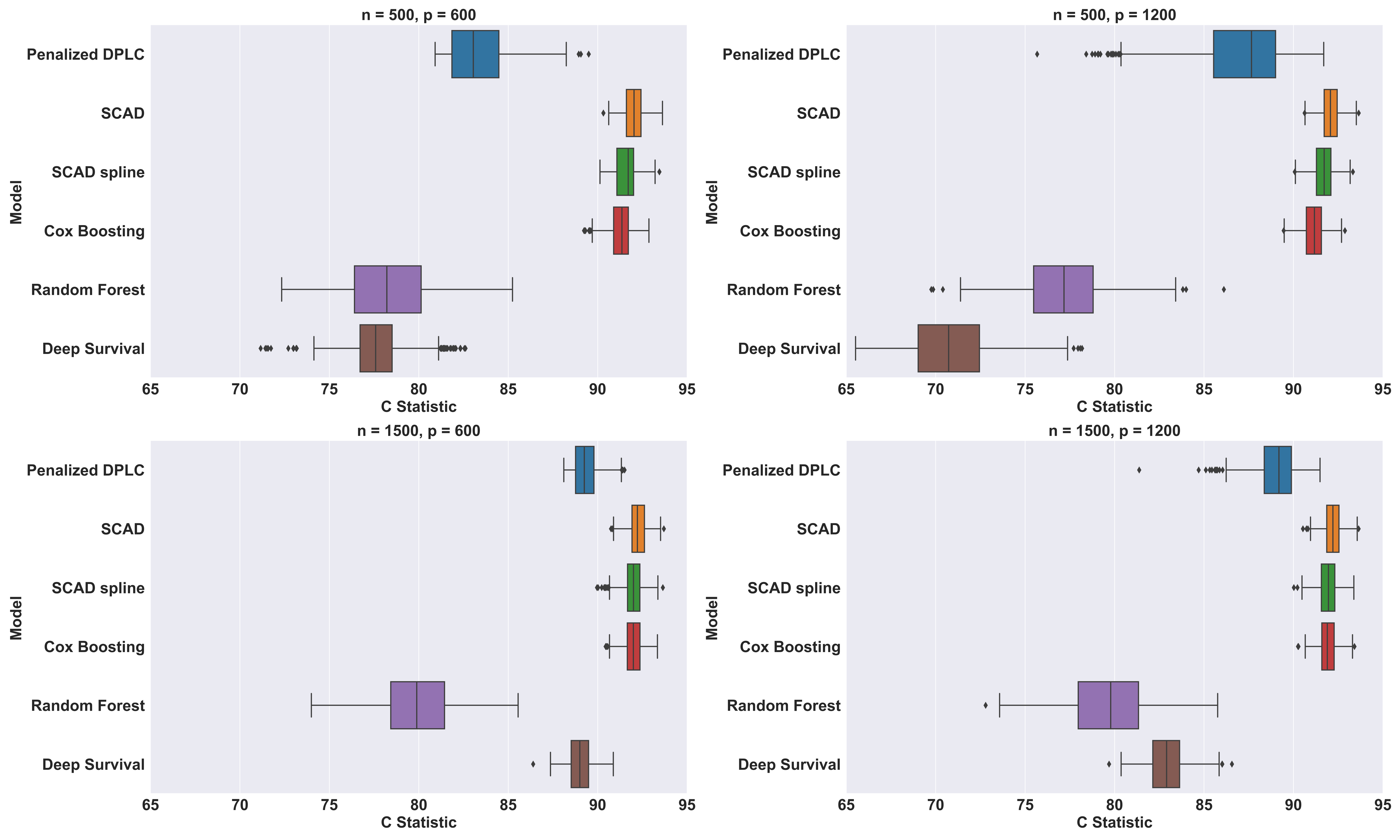}}
    \newline
    \subfloat[Nonlinear Case]{\includegraphics[scale =0.21]{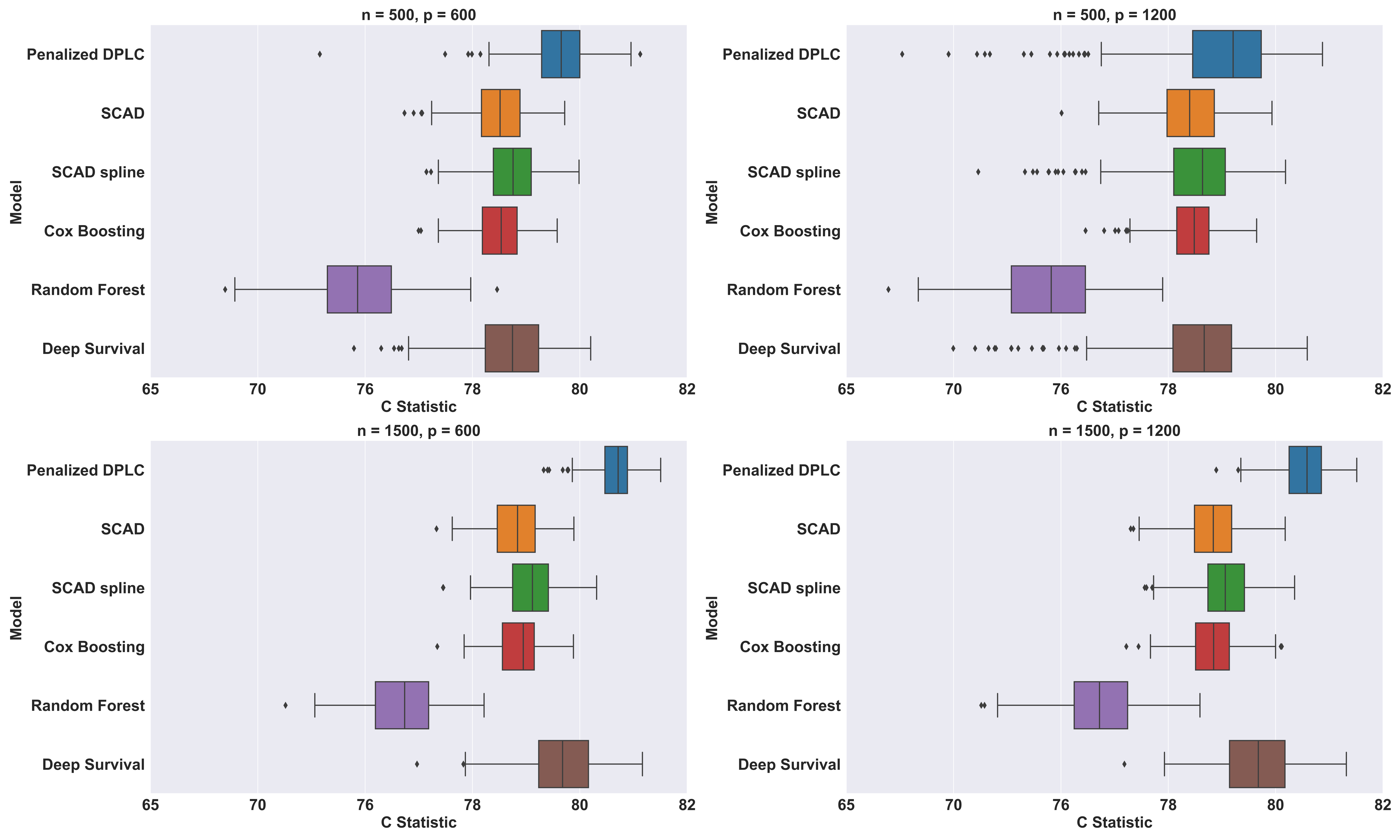}}
    \caption{\textbf{Prediction Performance Based on 500  Simulated Datasets} 
    }
    \label{sim c stat}
\end{figure}

Figure \ref{sim c stat} compared  the Penalized DPLC's prediction performance with the competing methods using the C-Index as the criterion. When $g_0$ is linear or the proportional hazards assumption holds, three Cox model-based methods, Cox-SCAD, SCAD splines, and Boosting, excelled with a highest median C-Index of approximately 0.92 across various combinations of $n$ and $p$. Our penalized DPLC yielded a competitive median C-Index, ranging from 0.83 to 0.87 at $n=500$ and improved to 0.89 at $n=1,500$; importantly, it outperformed two nonparametric methods: Random Forest (median C-Index values: 0.77--0.80) and Deep Survival Model (0.70--0.89).
 When $g_0$ is nonlinear, our Penalized DPLC model clearly outperformed the others  across various $n$ and $p$. The highest median C-Index of  0.868 [Interquartile range (IQR): 0.011] was achieved with $(n,p)=(1,500, 600)$. As the feature size increased, the prediction performance decreased slightly, e.g., the median C-Index for Penalized DPLC decreased from 0.841 (IQR: 0.018) to 0.830 (IQR: 0.032) when the feature size increased from 600 to 1,200 with 500 samples. The prediction performance improved with more samples; the median C-Index for Penalized DPLC rose to 0.865 (IQR: 0.015) when the sample size increased to 1,500, compared to 500 samples with 1,200 features.
  
To evaluate the selection performance, we reported the number of selected features, false positive number (FPN), false positive rate (FPR), false negative number (FNN), and false negative rate (FNR). Let $\mathcal{S}$ and $\widehat{\mathcal{S}}$ represent the actual and  estimated (i.e., the selected features) support of $\boldsymbol{\beta}$, and $Card(\cdot)$ the cardinality of a set.  Then
$   {\rm FPN} = Card(\widehat{\mathcal{S}} \backslash \mathcal{S}), $
   $ {\rm FPR} = {\rm FPN} /\{p - Card(\mathcal{S})\}$,
    ${\rm FNN} = Card( \mathcal{S} \backslash \widehat{\mathcal{S}}),$ and $
    {\rm FNR} = {\rm FNN} / Card(\mathcal{S}).$
When $g_0(\mathbf{z})$ is linear on $\mathbf{z}$, in which case the model assumptions were satisfied for the SCAD-penalized Cox model  with and without polynomial splines, they outperformed the other competing methods, including the Penalized DPLC (Table \ref{simulate select performance}). However, the performance of the penalized DPLC was comparable to them.
For example,  the two penalized Cox models 
 reported an FPN of less than 1,
while the penalized DPLC reported only 0.6 more FPNs on average than them. In addition, the average FNN, i.e., the missed `active' features, of the Penalized DPLC was only  0.74--2.31 (across various considered scenarios) higher than the penalized Cox models. On the other hand,  the performance of the Penalized DPLC was clearly better than  Cox Boosting and  Random Forest. Cox Boosting tended to select more features; when $(n,p)=(1,500,1,200)$,  Cox Boosting reported an FPN of 33.64 (SE: 0.60), whereas the FPN for the Penalized DPLC was 1.40 (SE: 0.10). For Random Forest, the average FNN varied from 4.12--6.17, compared to  1.33--3.49 for the Penalized DPLC. The average FPN for  Random Forest varied from 3.75--6.11, while it was  0.31--1.69 for the Penalized DPLC.
 
When $g_0(\mathbf{z})$ is nonlinear, 
the Penalized DPLC outperformed almost all of the other methods (except for Cox Boosting) in FNN. Cox Boosting had an FNN of 1.69 (SE: 0.04), while the Penalized DPLC reported a comparable FNN of 2.14 (SE: 0.04) with 500 samples and 1,200 features. However, Cox Boosting had a much higher FPN of 26.59 (SE: 0.47) compared to Penalized DPLC's  2.88 (SE: 0.12). The average number of falsely selected features using Penalized DPLC was 0.26--2.88, compared to 0.47--5.25 for the penalized Cox model. The selection performance of Penalized DPLC improved with more samples and fewer features, achieving the best performance when $(n,p)=(1,500,600)$  with an FPR of 0.04\% and an FNR of 11.16\%.

\begin{table}[hbpt]
    \begin{threeparttable}
    \centering
    \fontsize{9}{11}\selectfont
    \makebox[\textwidth]{\begin{tabular}{llccccc}
        \hline
        & \textbf{Method} &  \textbf{Selected Features}$^1$ & \textbf{FPN}$^2$ &          \textbf{FPR (\%)}$^3$ & \textbf{FNN}$^4$ &           \textbf{FNR(\%)}$^5$ \\
        \hline
        \textbf{Linear Case} & & & & & & \\
            \multirow{5}{*}{\textbf{(n,p) = (500,600)}} & Penalized DPLC &       6.90 (0.07) &    0.39 (0.04) &  0.07 (0.01) &    3.49 (0.04) &  34.90 (0.39) \\
                   & SCAD           &       9.47 (0.19) &    0.48 (0.09) &  0.08 (0.01) &    1.32 (0.11) &  13.20 (1.10) \\
                   & SCAD spline    &       9.51 (0.21) &    0.69 (0.15) &  0.12 (0.03) &    1.18 (0.11) &  11.80 (1.10) \\
                   & Cox Boosting   &      48.10 (1.02) &   38.75 (1.01) &  6.57 (0.17) &    0.65 (0.08) &   6.50 (0.82) \\
                   & Random Forest  &       9.51 (0.21) &    4.90 (0.26) &  0.83 (0.04) &    5.39 (0.13) &  53.90 (1.29) \\
                                \hline
            \multirow{5}{*}{\textbf{(n,p) = (500, 1,200)}} & Penalized DPLC &       9.90 (0.08) &    1.69 (0.09) &  0.14 (0.01) &    1.79 (0.05) &  17.86 (0.49) \\
                   & SCAD &       9.92 (0.08) &    0.85 (0.06) &  0.07 (0.01) &    0.93 (0.04) &   9.32 (0.40) \\
                   & SCAD spline &       9.94 (0.08) &    0.89 (0.07) &  0.07 (0.01) &    0.95 (0.04) &   9.48 (0.40) \\
                   & Cox Boosting &      49.22 (0.54) &   39.91 (0.54) &  3.35 (0.05) &    0.69 (0.03) &   6.88 (0.34) \\
                   & Random Forest &       9.94 (0.08) &    6.11 (0.10) &  0.51 (0.01) &    6.17 (0.06) &  61.74 (0.56) \\
            \hline
            \multirow{5}{*}{\textbf{(n,p) = (1,500, 600)}}& Penalized DPLC  &       8.99 (0.08) &    0.31 (0.07) &  0.05 (0.01) &    1.33 (0.04) &  13.26 (0.38) \\
                   & SCAD &       9.34 (0.03) &    0.01 (0.00) &  0.00 (0.00) &    0.67 (0.03) &   6.68 (0.33) \\
                   & SCAD spline &       9.63 (0.04) &    0.22 (0.03) &  0.04 (0.00) &    0.59 (0.03) &   5.88 (0.33) \\
                   & Cox Boosting &      42.68 (0.53) &   33.00 (0.53) &  5.59 (0.09) &    0.33 (0.02) &   3.26 (0.25) \\
                   & Random Forest &       9.63 (0.04) &    3.75 (0.07) &  0.64 (0.01) &    4.12 (0.06) &  41.20 (0.63) \\
                   \hline
            \multirow{5}{*}{\textbf{(n,p) = (1,500, 1,200)}} & Penalized DPLC &       9.87 (0.12) &    1.40 (0.10) &  0.12 (0.01) &    1.53 (0.05) &  15.32 (0.55) \\
                   & SCAD &       9.23 (0.04) &    0.04 (0.01) &  0.00 (0.00) &    0.81 (0.04) &   8.08 (0.39) \\
                   & SCAD spline &       9.65 (0.05) &    0.37 (0.04) &  0.03 (0.00) &    0.72 (0.04) &   7.16 (0.37) \\
                   & Cox Boosting &      43.19 (0.61) &   33.64 (0.60) &  2.83 (0.05) &    0.45 (0.03) &   4.46 (0.28) \\
                   & Random Forest &       9.65 (0.05) &    4.36 (0.08) &  0.37 (0.01) &    4.71 (0.06) &  47.08 (0.65) \\     
        \hhline{=======}
        \textbf{Nonlinear Case} & & & & & & \\
            \multirow{5}{*}{\textbf{(n,p) = (500, 600)}} & Penalized DPLC &      11.04 (0.11) &    2.52 (0.10) &  0.43 (0.02) &    1.48 (0.03) &  14.76 (0.29) \\
                   & SCAD &      12.68 (0.19) &    4.66 (0.17) &  0.79 (0.03) &    1.98 (0.05) &  19.78 (0.47) \\
                   & SCAD spline &      12.32 (0.18) &    4.13 (0.16) &  0.70 (0.03) &    1.81 (0.05) &  18.08 (0.46) \\
                   & Cox Boosting &      34.73 (0.49) &   26.02 (0.48) &  4.41 (0.08) &    1.29 (0.04) &  12.90 (0.38) \\
                   & Random Forest &      12.32 (0.18) &    7.27 (0.18) &  1.23 (0.03) &    4.95 (0.06) &  49.50 (0.59) \\
            \hline
            \multirow{5}{*}{\textbf{(n,p) = (500, 1,200)}} & Penalized DPLC  &      10.74 (0.13) &    2.88 (0.12) &  0.24 (0.01) &    2.14 (0.04) &  21.38 (0.37) \\
                   & SCAD &      12.89 (0.21) &    5.25 (0.19) &  0.44 (0.02) &    2.36 (0.05) &  23.64 (0.51) \\
                   & SCAD spline &      17.87 (0.97) &   10.02 (0.97) &  0.84 (0.08) &    2.15 (0.05) &  21.48 (0.48) \\
                   & Cox Boosting &      34.90 (0.49) &   26.59 (0.47) &  2.23 (0.04) &    1.69 (0.04) &  16.94 (0.44) \\
                   & Random Forest &      17.87 (0.97) &   13.16 (0.96) &  1.11 (0.08) &    5.28 (0.05) &  52.84 (0.51) \\
            \hline
            \multirow{5}{*}{\textbf{(n,p) = (1,500, 600)}}& Penalized DPLC &       9.14 (0.04) &    0.26 (0.02) &  0.04 (0.00) &    1.12 (0.03) &  11.16 (0.33) \\
                   & SCAD &       8.76 (0.05) &    0.47 (0.03) &  0.08 (0.01) &    1.71 (0.05) &  17.06 (0.45) \\
                   & SCAD spline &      10.49 (0.11) &    1.82 (0.10) &  0.31 (0.02) &    1.32 (0.04) &  13.24 (0.40) \\
                   & Cox Boosting &      33.12 (0.52) &   24.02 (0.51) &  4.07 (0.09) &    0.90 (0.03) &   9.00 (0.34) \\
                   & Random Forest &      10.49 (0.11) &    4.08 (0.13) &  0.69 (0.02) &    3.58 (0.06) &  35.84 (0.59) \\
            \hline
            \multirow{5}{*}{\textbf{(n,p) = (1,500, 1,200)}} & Penalized DPLC &       9.20 (0.08) &    0.94 (0.06) &  0.08 (0.00) &    1.74 (0.05) &  17.40 (0.46) \\
                   & SCAD &       9.04 (0.09) &    1.16 (0.07) &  0.10 (0.01) &    2.12 (0.05) &  21.20 (0.50) \\
                   & SCAD spline &      10.35 (0.13) &    2.17 (0.11) &  0.18 (0.01) &    1.83 (0.05) &  18.26 (0.47) \\
                   & Cox Boosting &      33.20 (0.55) &   24.56 (0.54) &  2.06 (0.05) &    1.36 (0.04) &  13.58 (0.40) \\
                   & Random Forest &      10.35 (0.13) &    4.60 (0.13) &  0.39 (0.01) &    4.26 (0.06) &  42.56 (0.61) \\
         \hline
    \end{tabular}}
    \begin{tablenotes}\footnotesize
    \item[\textsuperscript{1}]The  number of  true `active' features is set to be ten.
    \item[\textsuperscript{2}] False Positive Number (FPN) is the number of features that are `inactive' but selected by the model as `active' features. 
    \item[\textsuperscript{3}] False Positive Rate (FPR) is the FPN divided by the true number of `inactive' features and reported as a percentage ($\times 100$).
    \item[\textsuperscript{4}] False Negative Number (FNN) is the number of features that are `active' but selected by the model as `inactive' features.
    \item[\textsuperscript{5}] False Negative Number (FNR) is the FNN divided by the true number of `active' features and reported as a percentage ($\times 100$).
    \item[*] Reported numbers are means and standard errors (SEs).
    \end{tablenotes}
    \caption{\textbf{Selection Performance of Different Algorithms using 500 Simulated Datasets}}
    \label{simulate select performance}
    \end{threeparttable}
\end{table}

\section{Application}
We applied the Penalized DPLC to analyze a dataset from  NLST,  investigating what and how  CT features were related to the mortality of lung cancer patients. The dataset includes a total of 368 subjects from NLST who were diagnosed with lung cancer and screened with  CT (Table \ref{descriptive analysis}). Out of them,  96 patients died  during follow-up. The median age  was 63.5 years old (IQR: 59.0, 68.0), with  55\% being male  and over 90\% being white. 
Most patients were in the early cancer stage, and hypertension was the most prevalent comorbidity (36\%), followed by obstructive lung disease (24\%) and prior pneumonia (21\%).

To extract features from CT scans, we followed the image processing pipeline as outlined in Figure S.2 in the Supplement Material. We first removed noise from the images through gray-scale normalization and adaptive histogram equalization. We then normalized the voxel intensity of each image to a standard range of 0 (black) to 255 (white) units and improved the contrast with adaptive histogram equalization. We next identified the regions of interest (ROIs) and segmented the tumor regions based on their location and size. We used \textit{pyradiomics} to extract texture features from the ROIs,
including first-order features, shape-based features, and higher-order features~\citep{amadasun1989textural}. We applied image filtration using the Laplacian of Gaussian filter and a 3D LBP-based filter; the Laplacian of Gaussian filter highlights areas of gray level change~\citep{kong2013Generalized}, and the 3D LBP-based filter computes local binary patterns in 3D using spherical harmonics~\citep{banerjee20123d}. A total of 320 image features were extracted.

To compare the prediction and selection accuracy of the Penalized DPLC with other competing methods, we conducted 100 experiments. In each experiment, we tuned the number of hidden layers and the number of neurons in each hidden layer over the grids of [1, 2, 3, 4] and [2, 4, 8, 16], respectively, when constructing the DNN, 
and randomly divided the data into 80\% for training and the remaining 20\% for testing. To ensure that the censoring rate in the training and testing data remained the same as in the entire population, we split the data  by stratifying the vital status of the patients. Similar to the simulation study, we tuned the number of hidden layers and the number of neurons in each hidden layers over the grid of [1, 2, 3, 4] and [2, 4, 8, 16], respectively.

\begin{figure}
    \centering
    \includegraphics[scale =0.3]{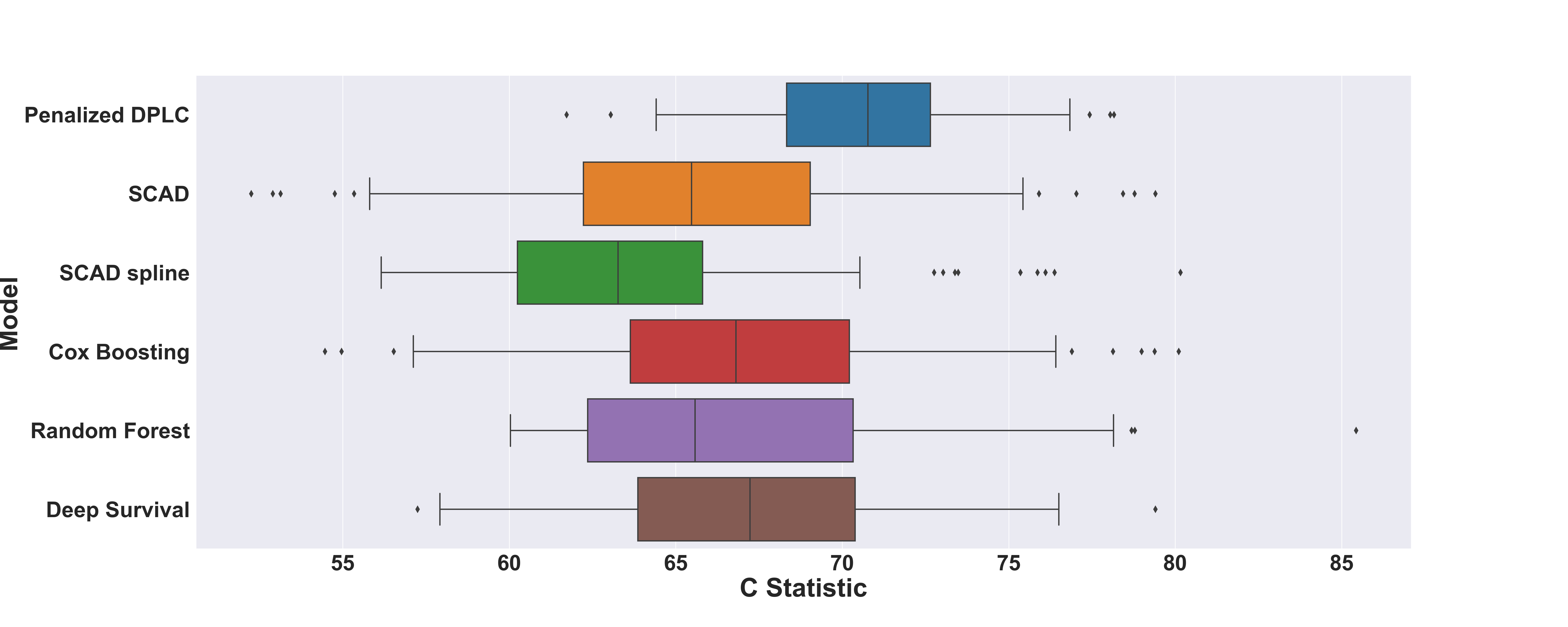}
    \caption{\textbf{Prediction Performance of 100 Experiments using Data from the National Lung Cancer Screen Trial:} During each experiment, 80\% data is randomly selected as training data, and 20\% data is selected as testing data. The censoring rate in the testing data and training data is controlled to be the same as that in the entire population.}
    \label{real c stat}
\end{figure}

As shown in Figure \ref{real c stat}, the median C-Index for Penalized DPLC is 0.708 (IQR: 0.043), outperforming the other competing methods. Deep Survival (Median: 0.672, IQR: 0.065), Random Forest (Median: 0.656, IQR: 0.080), and Cox Boosting (Median: 0.668, IQR: 0.066) all had better prediction performance than the SCAD-penalized Cox model (Median: 0.655, IQR: 0.068) and the SCAD-penalized partially linear Cox model (Median: 0.633, IQR: 0.065).

\begin{figure}[ht]
    \centering
    \subfloat[Gradient for $\widehat{g}$ of age]{\label{gradient_age}{\includegraphics[width=0.35\textwidth]{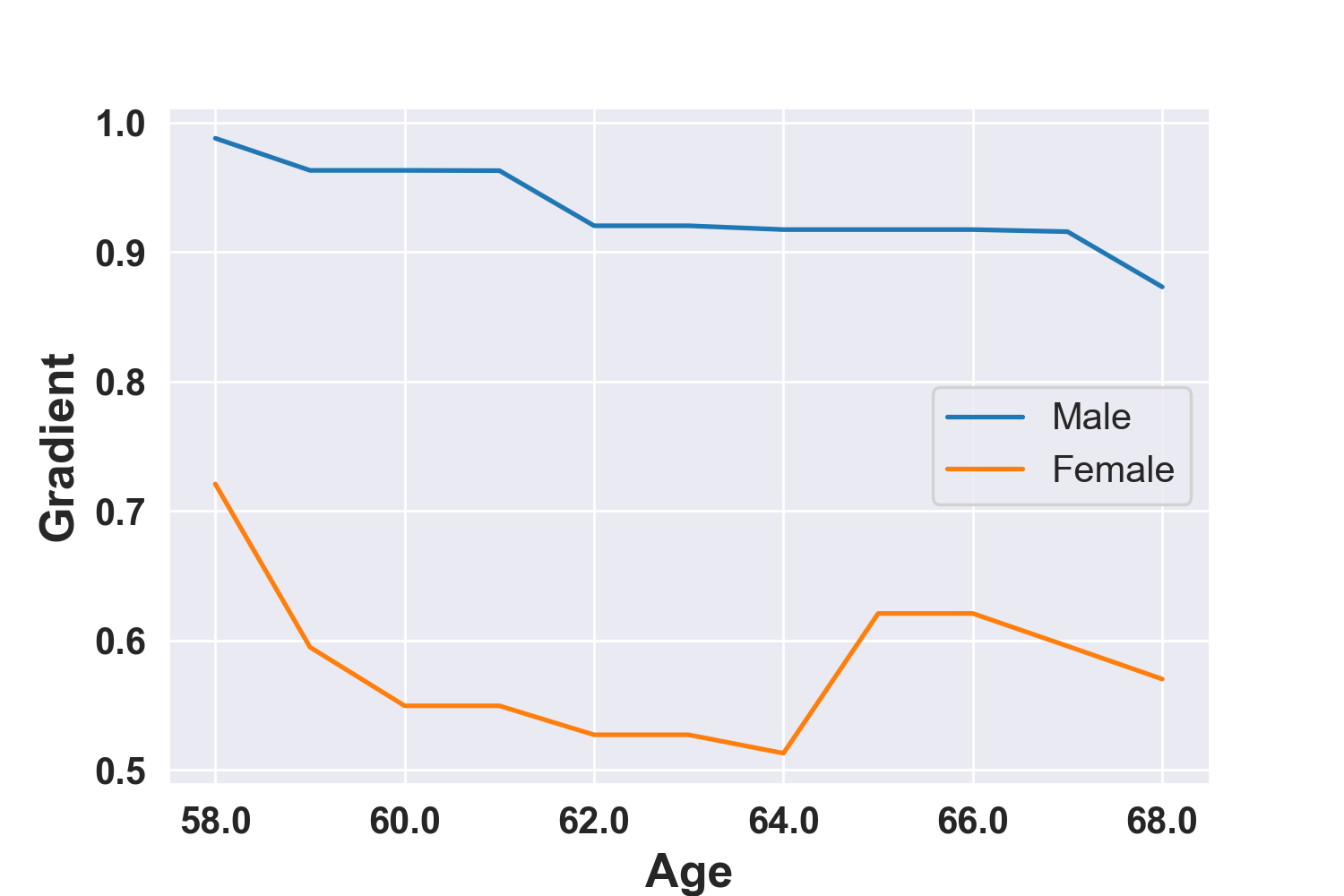}}}\hspace{-1.5em}
    \subfloat[Gradient for $\widehat{g}$ of BMI]{\label{gradient_bmi}{\includegraphics[width=0.35\textwidth]{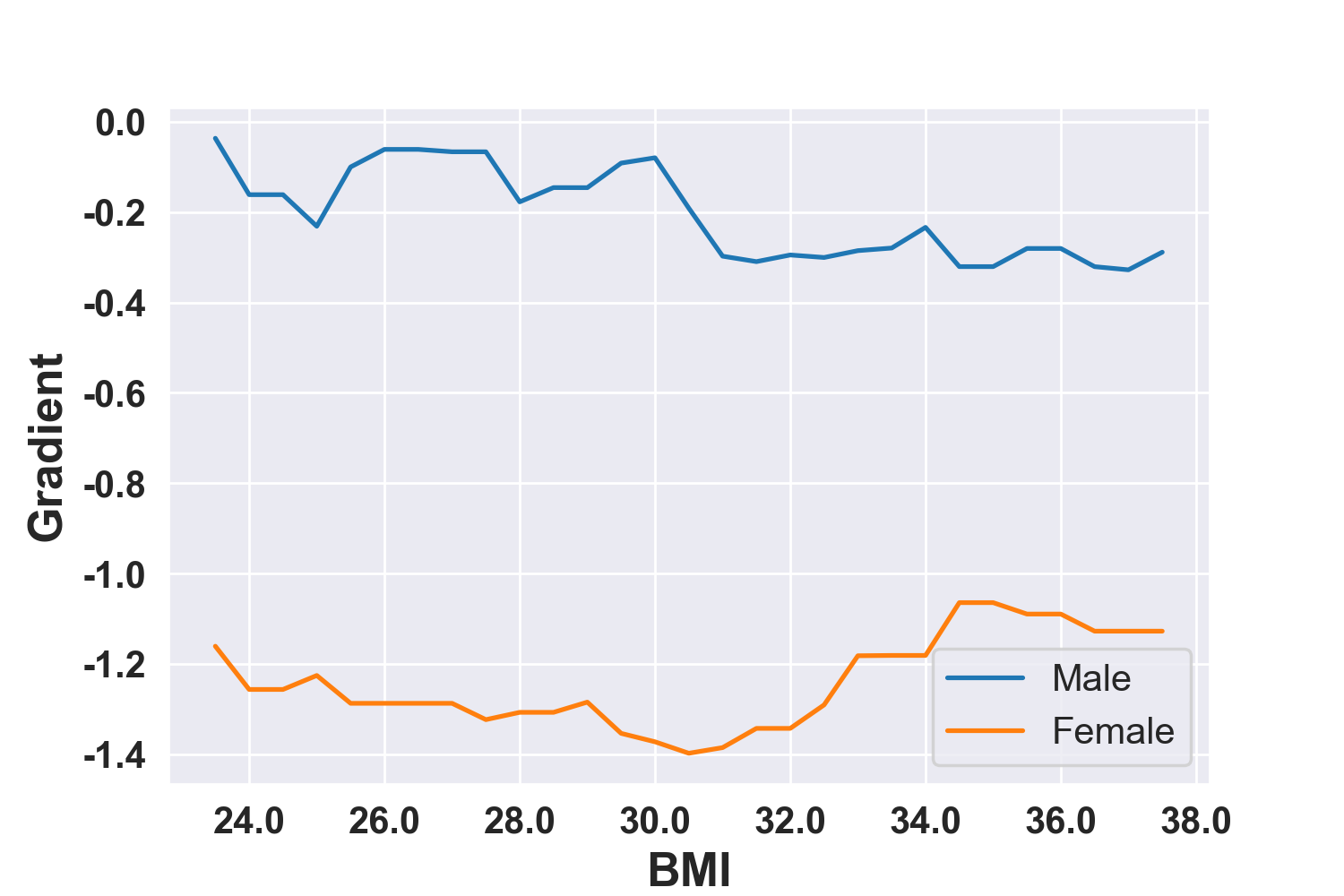}}}\hspace{-1.5em}
    \subfloat[Gradient for $\widehat{g}$ of pack years of smoking]{\label{gradient_pack_year}{\includegraphics[width=0.35\textwidth]{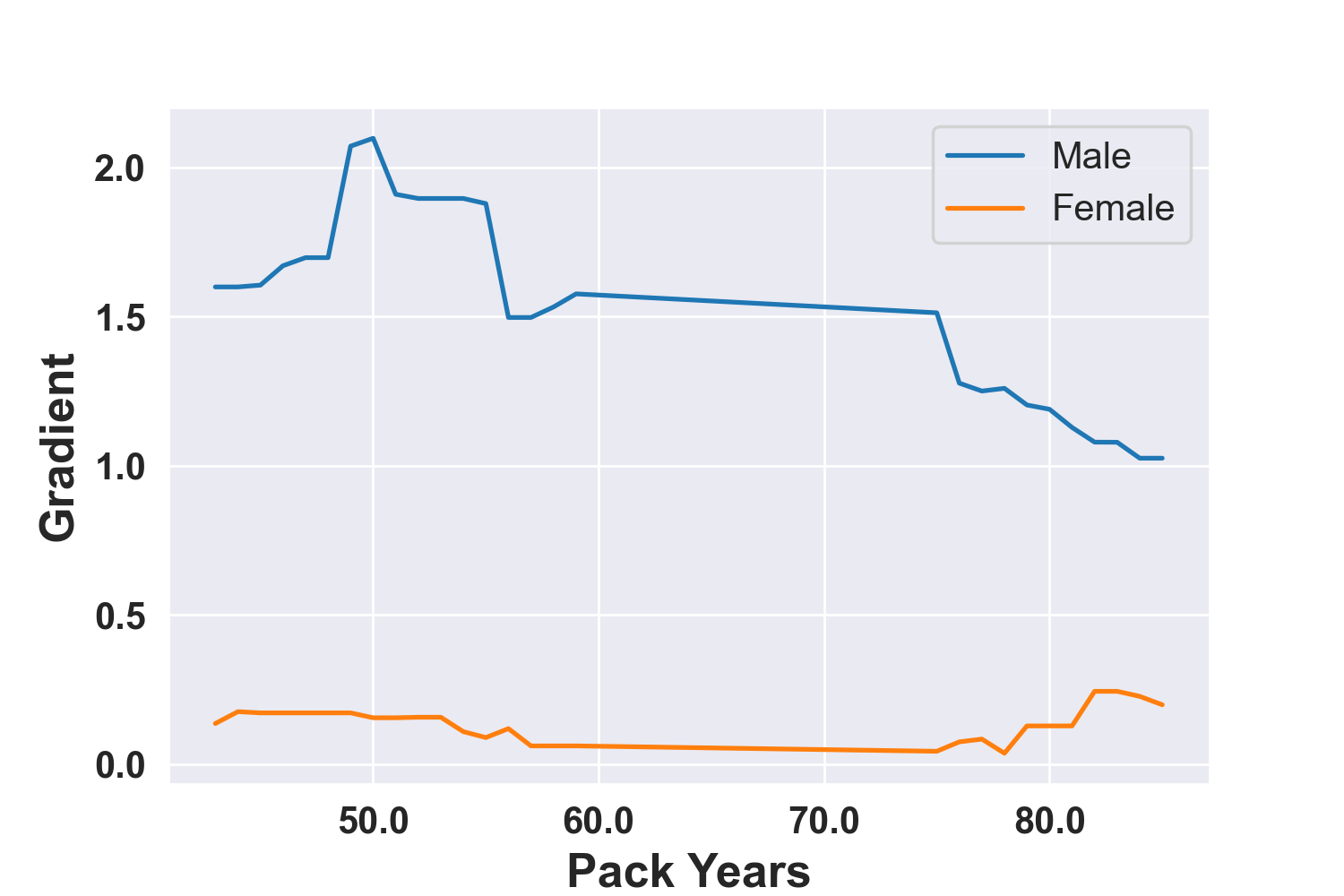}}}\hspace{-1.5em}
    \subfloat[$\widehat{g}$ of age and BMI]{\label{g_age_bmi}{\includegraphics[width=0.35\textwidth]{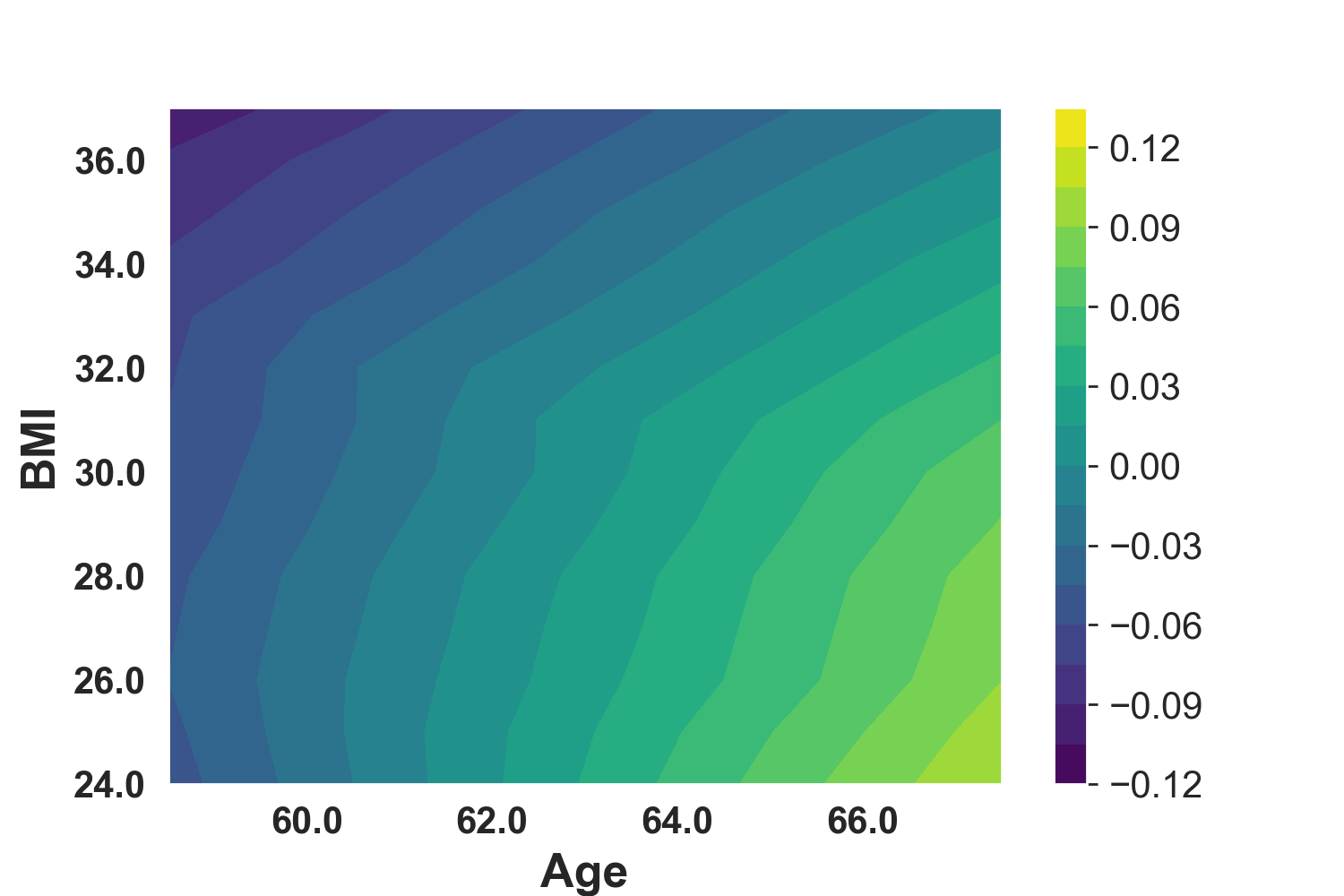}}}\hspace{-1.5em}
    \subfloat[$\widehat{g}$ of pack years of smoking  and BMI]{\label{g_pack_year_bmi}{\includegraphics[width=0.35\textwidth]{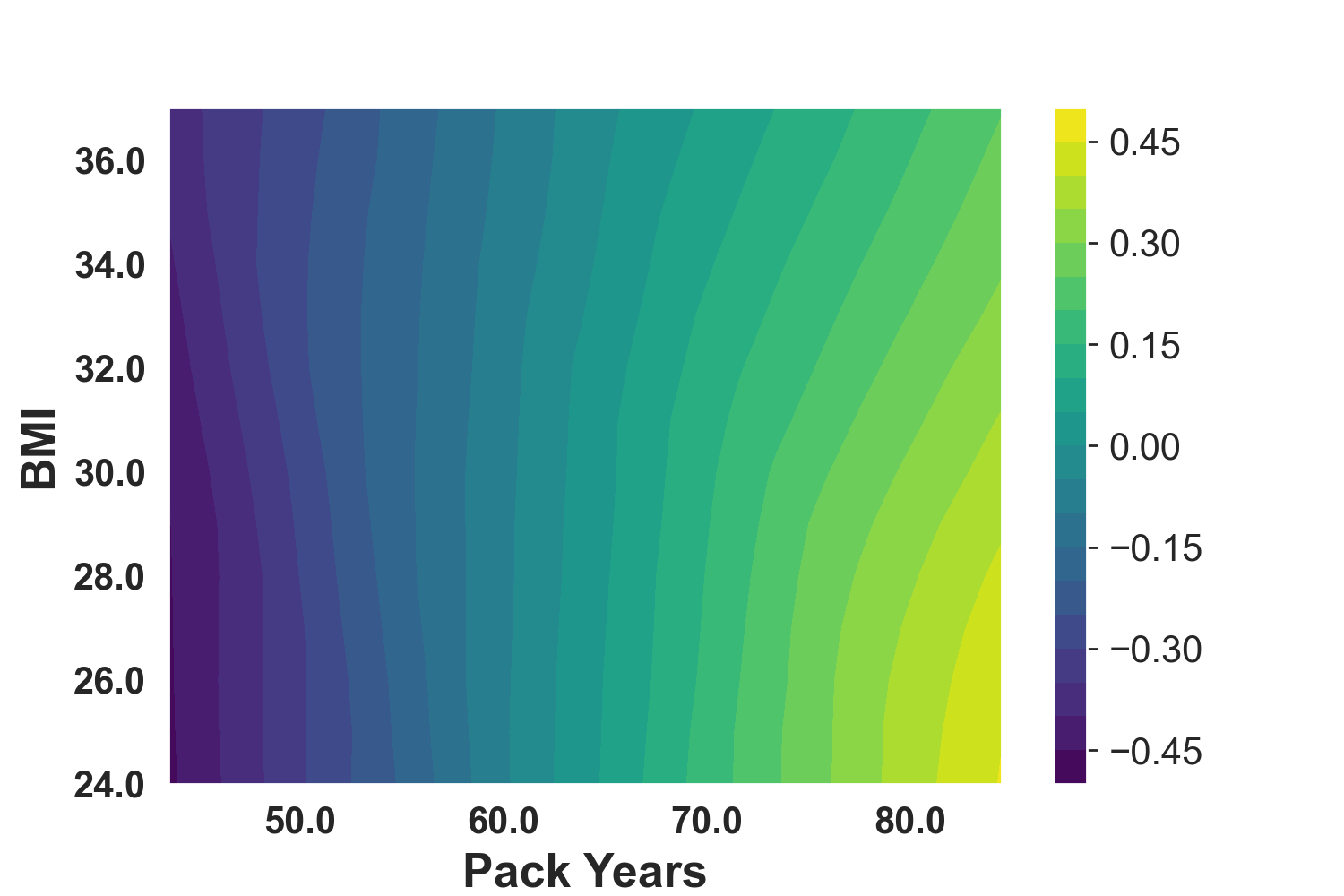}}}
    \subfloat[$\widehat{g}$ of age  and pack years of smoking]{\label{g_age_pack_year}\hspace{-1.5em}{\includegraphics[width=0.35\textwidth]{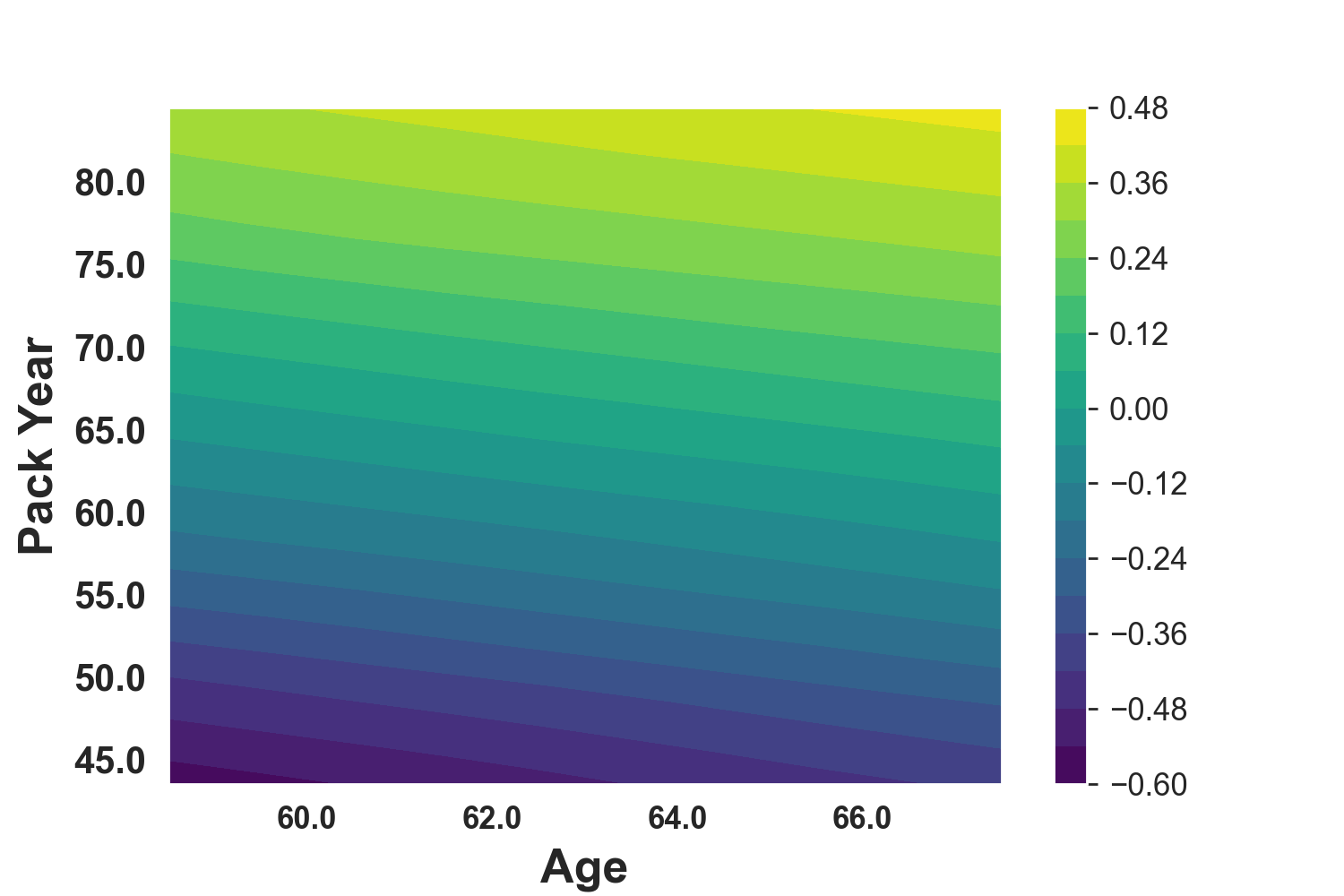}}}
     \caption{\textbf{Estimated Nonlinear Function and Gradients using NLST:} The gradients for $\widehat{g}$ of age, BMI, and pack years smoking history stratified by gender are plotted in (a), (b), and (c). $\widehat{g}$ of age, BMI, and pack years of smoking is plotted in (d) and (e). The other variables are fixed at their sample means (for continuous variables) or modes (for categorical variables)}
\end{figure}

 Figures \ref{g_age_bmi}--\ref{g_age_pack_year} illustrate the estimated effects of age, BMI, and pack years of smoking  while holding other variables constant at their mean (for continuous variables) or mode (for categorical variables), as derived from the estimated $\widehat{g}$ function.
These contour plots clearly reveal the nonlinear relationships between age, BMI, and pack years of smoking  and survival. The gradients of $\widehat{g}$ for age, BMI, and pack years, stratified by gender, are presented in Figures \ref{gradient_age}--\ref{gradient_pack_year}, reflecting the local change in the log hazard for small changes in the corresponding variables. Figures \ref{gradient_age} and \ref{gradient_pack_year} exhibit positive gradients for age and pack years, indicating that mortality increases with increasing age and pack years, consistent with the literature \citep{tindle2018lifetime}. In contrast, Figure \ref{gradient_bmi} shows that BMI has a protective effect on patient survival, in agreement with the obesity paradox \citep{lee2019obesity}.
Moreover, we observe that gender has a significant impact on lung cancer survival. As seen in the gradient figures, male patients exhibit a steeper increase in mortality risk compared to female patients for small increments in age and pack years, as shown in Figures \ref{gradient_age} and \ref{gradient_pack_year}. On the other hand, Figure \ref{gradient_bmi} highlights that an increased BMI has a stronger protective effect for female patients compared to male patients,
consistent with previous findings of better survival outcomes for female patients 
 \citep{visbal2004gender}.


The Penalized DPLC method has selected five radiomic features as risk factors: large dependence low gray level emphasis (LDLGLE), large area emphasis (LAE), large area low gray level emphasis (LALGLE), cluster shade, and contrast. Figure S.3 in the Supplement Material demonstrates the reproducibility of feature selection by the Penalized DPLC and the hazard ratios for the selected features. LDLGLE (HR: 1.07) and cluster shade (HR: 1.09) were selected 71 and 57 times out of 100 experiments, respectively. Although LALGLE (Frequency: 51, HR: 1.02) and contrast (Frequency: 41, HR: 1.02) were selected less frequently than the other texture features, they were still more frequently selected by the Penalized DPLC.

The selected radiomic features  have biological significance.  LDLGLE and LALGLE represent the extent of low voxel intensities or soft-tissue attenuation, indicating the presence of lymphatic or vascular invasion~\citep{higgins2012lymphovascular};
LAE, cluster shade, and contrast quantify the roughness and heterogeneity of textures~\citep{amadasun1989textural}. 

\begin{table}[hbpt]
    \begin{threeparttable}
    \centering
    \fontsize{9}{11}\selectfont
    \makebox[\textwidth]{\begin{tabular}{lccc}
        \hline
        \textbf{Characteristic} & \textbf{Overall}, N = 368\textsuperscript{1} & \textbf{Alive}, N = 272\textsuperscript{1} & \textbf{Dead}, N = 96\textsuperscript{1} \\ 
        \hline
        Median Follow-up Time (days) & 2072 (1962, 2151) &  \\
        Age (yrs.) & 63.5 (59.0, 68.0) & 63.0 (59.0, 67.0) & 66.0 (60.0, 70.0) \\ 
        BMI & 26.3 (24.3, 29.2) & 26.3 (24.3, 29.2) & 26.1 (24.1, 29.2) \\
        Gender &  &  &  \\ 
            \qquad Male & 201 (55\%) & 137 (50\%) & 64 (67\%) \\ 
            \qquad Female & 167 (45\%) & 135 (50\%) & 32 (33\%) \\ 
        Race &  &  &  \\ 
            \qquad White & 339 (92\%) & 251 (92\%) & 88 (92\%) \\ 
            \qquad Black & 14 (3.8\%) & 11 (4.0\%) & 3 (3.1\%) \\ 
            \qquad Asian & 8 (2.2\%) & 6 (2.2\%) & 2 (2.1\%) \\ 
            \qquad Other & 6 (1.6\%) & 3 (1.1\%) & 3 (3.1\%) \\ 
            \qquad Unknow & 1 (0.3\%) & 1 (0.4\%) & 0 (0\%) \\ 
        Cigarette Smoking Status &  &  &  \\ 
            \qquad Former & 171 (46\%) & 135 (50\%) & 36 (38\%) \\ 
            \qquad Current & 197 (54\%) & 137 (50\%) & 60 (62\%) \\ 
        Pack Years of Smoking  & 58 (46, 80) & 57 (45, 80) & 60 (49, 84) \\ 
        Histology &  &  & \\ 
            \qquad Adenocarcinoma & 185 (50\%) & 137 (50\%) & 48 (50\%) \\ 
            \qquad Squamous Cell Carcinoma & 73 (20\%) & 50 (18\%) & 23 (24\%) \\ 
            \qquad Large Cell Carcinoma & 16 (4.3\%) & 9 (3.3\%) & 7 (7.3\%) \\ 
            \qquad Adenosquamous Carcinoma & 8 (2.2\%) & 3 (1.1\%) & 5 (5.2\%) \\ 
            \qquad Neuroendocrine/Carcinoid Tumors & 1 (0.3\%) & 1 (0.4\%) & 0 (0\%) \\ 
            \qquad Bronchioloalveolar Carcinoma & 70 (19\%) & 59 (22\%) & 11 (11\%) \\ 
            \qquad NSCLC NOS & 15 (4.1\%) & 13 (4.8\%) & 2 (2.1\%) \\ 
        Pathologic Stage &  &  &  \\ 
            \qquad IA & 230 (62\%) & 188 (69\%) & 42 (44\%) \\ 
            \qquad IB & 49 (13\%) & 36 (13\%) & 13 (14\%) \\ 
            \qquad IIA & 11 (3.0\%) & 8 (2.9\%) & 3 (3.1\%) \\ 
            \qquad IIB & 39 (11\%) & 26 (9.6\%) & 13 (14\%) \\ 
            \qquad  IIIA & 33 (9.0\%) & 13 (4.8\%) & 20 (21\%) \\ 
            \qquad IIIB & 3 (0.8\%) & 1 (0.4\%) & 2 (2.1\%) \\ 
            \qquad IV & 3 (0.8\%) & 0 (0\%) & 3 (3.1\%) \\ 
        Radiotherapy & 27 (7.3\%) & 9 (3.3\%) & 18 (19\%)\\ 
        Chemotherapy & 83 (23\%) & 49 (18\%) & 34 (35\%)\\ 
        Surgery Type &  &  &  \\ 
            \qquad Wedge/Multiple Wedge Resection & 45 (12\%) & 30 (11\%) & 15 (16\%) \\ 
            \qquad Segmentectomy & 14 (3.8\%) & 8 (2.9\%) & 6 (6.2\%) \\ 
            \qquad Lobectomy & 287 (78\%) & 222 (82\%) & 65 (68\%) \\ 
            \qquad Bilobectomy & 15 (4.1\%) & 9 (3.3\%) & 6 (6.2\%) \\ 
            \qquad Pneumonectomy & 7 (1.9\%) & 3 (1.1\%) & 4 (4.2\%) \\ 
        Asthma & 27 (7.3\%) & 18 (6.6\%) & 9 (9.4\%) \\ 
        Bronchitis & 35 (9.5\%) & 23 (8.5\%) & 12 (12\%) \\ 
        COPD & 39 (11\%) & 24 (8.8\%) & 15 (16\%) \\ 
        Diabetes & 33 (9.0\%) & 20 (7.4\%) & 13 (14\%) \\ 
        Emphysema & 48 (13\%) & 32 (12\%) & 16 (17\%) \\ 
        Heart Disease & 52 (14\%) & 35 (13\%) & 17 (18\%) \\ 
        Hypertension & 134 (36\%) & 98 (36\%) & 36 (38\%) \\ 
        Prior Pneumonia & 77 (21\%) & 53 (19\%) & 24 (25\%) \\ 
        Obstructive Lung Disease & 88 (24\%) & 58 (21\%) & 30 (31\%) \\ 
      \hline
    \end{tabular}}
    \begin{tablenotes}\footnotesize
    \item[\textsuperscript{1}]Median (IQR); n (\%)
    \end{tablenotes}
    \caption{\textbf{Clinical Characteristics of Patients from the National Lung Cancer Screen Trial}}
    \label{descriptive analysis}
    \end{threeparttable}
\end{table}



\section{Discussion}

To address the analytical needs of the National Lung Screening Trial (NLST), we propose the Penalized DPLC model, which simultaneously selects and models the effects of prognostic radiomic features. 
Our adopted partial linear model assumes a log-linear relationship between radiomic features and hazards, allowing us to use the SCAD penalty to identify important image features. 
Clinical features with known associations with survival outcomes are modeled using a nonparametric function to account for their nonlinear effects. Despite this structured approach, we maintain the flexibility to model selected radiomic features using nonparametric functions like the clinical features. Our method provides a convenient means
to explore  new predictors while fully characterizing the impact of established risk factors.



There is significant potential for future work. Our modeling framework can be extended to incorporate alternative penalties, such as the LASSO and MCP~\citep{tibshirani2011regression}.
We are currently utilizing a DNN estimator with a fixed and moderate dimension, which is suitable for our dataset where the number of clinical variables is moderate.  It is feasible to develop DNN estimators that can handle high-dimensional predictors. Moreover, quantifying the uncertainty of the estimates remains a significant challenge.

\newpage
\renewcommand\thefigure{\thesection\arabic{figure}}    
\setcounter{figure}{0}   
\renewcommand\thetable{\thesection\arabic{table}}    
\setcounter{table}{0}   
\begin{center}
    APPENDIX
\end{center}

\begin{appendix}
\section{Composite H\"{o}lder Class of Smooth Functions}

With constants $a, M >0$ and a positive integer $d$, we define a H\"{o}lder class of smooth functions  as 
\[
    \mathcal{H}_{d}^{a} (\mathbb{D}, M) = \{ f: \mathbb{D} \subset \mathbb{R}^d \to \mathbb{R}:\sum_{\upsilon:|\upsilon| < a} \Vert\partial^{\upsilon} f \Vert_{\infty} + \sum_{\upsilon:|\upsilon| = \lfloor a \rfloor} \sup_{x,y \in \mathbb{D},x\neq y} \frac{|\partial^{\upsilon}f(x) - \partial^{\upsilon}f(y) |}{\Vert x - y\Vert_{\infty}^{a - \lfloor a \rfloor}} \leq M \},
\]
where $\mathbb{D}$ is a bounded subset of $\mathbb{R}^d$, $\lfloor a \rfloor$ is the largest integer smaller than $a$, $\partial^{\upsilon} := \partial^{\upsilon_1}\dots\partial^{\upsilon_r}$ with $\upsilon = (\upsilon_1,\dots,\upsilon_d) \in \mathbb{N}^d$, and $|\upsilon| := \sum_{j=1}^d\upsilon_j$. 

For a positive integer $q$, 
let $\alpha = (\alpha_1, \dots, \alpha_q) \in \mathbb{R}_+^{q}$, and $\mathbf{d} = (d_1,\dots,d_{q+1})\in \mathbb{N}_+^{q+1}$, $\Tilde{\mathbf{d}} = (\Tilde{d}_1,\dots,\Tilde{d}_q) \in \mathbb{N}_+^{q}$ with $\Tilde{d}_j \leq d_j$. 
We then define  a composite H\"{o}lder  smooth function class as
\begin{equation}
    \mathcal{H}(q,\alpha,\mathbf{d},\Tilde{\mathbf{d}},M) = \{f = f_q \circ\dots\circ f_1: f_i = (f_{i1}, \dots, f_{id_{i+1}})^{\top}, f_{ij} \in \mathcal{H}_{\Tilde{d}_i}^{\alpha_i}([a_i,b_i]^{\Tilde{d}_i},M), |a_i|, |b_i| \leq M \},
\end{equation}
where $[a_i, b_i]$ is the bounded domain for each H\"{o}lder  smooth function.
\section{More Notation}
Denote $a_n \lesssim b_n$ as $a_b \leq c b_n$ for some $c > 0$ when $n$ is sufficiently large; $a_n \asymp b_n$ if $a_n \lesssim b_n$ and $b_n \lesssim a_n$. Let $\eta(\cdot, \cdot) = (\boldsymbol{\beta}^\top \cdot, g(\cdot)): \mathbb{R}^p
\times \mathbb{R}^r \to \mathbb{R}^2$  denote the collection of a linear operator and a nonlinear operator. In this section, denote by
$\mathbf{v} = (\mathbf{x}^\top,\mathbf{z}^\top)^\top$ the random vector underlying the observed
IID data of $\mathbf{v}_i=(\mathbf{x}_i^\top,\mathbf{z}_i^\top)^\top$,
and  $(T,\Delta)$  the random vector underlying the observed IID data of
$(T_i,\Delta_i), i=1, \ldots, n$. Let $N(t) = I(T\leq t, \Delta = 1)$ and $N_i(t) = I(T_i\leq t, \Delta_i = 1)$. To simplify notation, we denote by ${\eta}(\mathbf{v}) = \boldsymbol{\beta}^{\top}\mathbf{x} + g(\mathbf{z})$. Denote the truth of  $\eta(\cdot, \cdot)$ by $\eta_0(\cdot, \cdot)= (\boldsymbol{\beta_0}^\top\cdot, g_0(\cdot))$.
For two operators, say, $\eta_1(\cdot, \cdot)=(\boldsymbol{\beta}_1^\top\cdot, g_1(\cdot))$ and $\eta_2(\cdot, \cdot)=(\boldsymbol{\beta}_2^\top\cdot, g_2(\cdot))$, define their distance as  \[d^2(\eta_1,\eta_2) \defeq \mathbb{E}  [\{{\eta_1}(\mathbf{v}) - {\eta_2}(\mathbf{v})\}^2]=
\int \{ {\eta_1}(\mathbf{t}) -  {\eta_2}(\mathbf{t})\}^2 f_\mathbf{v}(\mathbf{t})
d \mathbf{t},
\]
and   the corresponding norm 
$$\Vert \eta \Vert^2 \defeq  \mathbb{E}  [ {\eta}^2(\mathbf{v})] =\int  {\eta}^2(\mathbf{t}) f_\mathbf{v}(\mathbf{t})
d \mathbf{t}.$$ For the notational ease, we write $\eta=(\boldsymbol{\beta}, g)$ in the following.

With $Y(t) = 1(T \geq t)$ and $Y_i(t) = 1(T_i \geq t)$,  define
\begin{align*}
    S_{0n}(t,\eta) &= \frac{1}{n}\sum_{i=1}^n Y_i(t)\exp\{  {\eta}(\mathbf{v}_i)\}, 
    \qquad 
    S_0(t,\eta) = \mathbb{E}[Y(t)\exp\{  {\eta}(\mathbf{v})\} ],
    \end{align*}
and   for any vector function $\mathbf{h}$ of $\mathbf{v}$ define
    \begin{align*}
    S_{1n}(t,\eta,\mathbf{h}) &= \frac{1}{n}\sum_{i=1}^n Y_i(t)\mathbf{h}(\mathbf{v}_i)\exp\{  {\eta}(\mathbf{v}_i)\},
    \qquad
    S_{1}(t,\eta,\mathbf{h}) = \mathbb{E}[Y(t)\mathbf{h}(\mathbf{v})\exp\{  {\eta}(\mathbf{v})\}],
\end{align*}
where the expectation is taken with respect to the joint distribution of $T$ and $\mathbf{v}$.

Let
\begin{align*}
    l_n(t,\mathbf{v},\eta) =  {\eta}(\mathbf{v}) - \log S_{0n}(t,\eta),
    \qquad 
    l_0(t,\mathbf{v},\eta) =  {\eta}(\mathbf{v}) - \log S_{0}(t,\eta).
\end{align*}
Then the partial likelihood in (2)  


can be written as
\begin{align*}
    \ell(\eta) = \frac{1}{n}\sum_{i=1}^n\{\Delta_i l_n(T_i,\mathbf{v}_i,\eta) -\Delta_i\log n\}.
\end{align*}
Since $\sum_{i=1}^n \Delta_i\log n$ does not involve unknown parameters and can be dropped in optimization, we replace below $\ell(\eta)$ by $\frac{1}{n}\sum_{i=1}^n\{\Delta_i l_n(T_i,\mathbf{v}_i,\eta)\}$.

Finally, for any function $h$ of $(\mathbf{v},\Delta,T)$, where $(\Delta,T)$
is the random vector underlying $(\Delta_i,T_i)$, define
\begin{align*}
    \mathbb{P}_n \{h(\mathbf{v},\Delta,T)\} = \frac{1}{n}\sum_{i=1}^n h(\mathbf{v}_i,\Delta_i,T_i),
    \qquad
    \mathbb{P}\{h(\mathbf{v},\Delta,T)\} = \mathbb{E} \{h(\mathbf{v},\Delta,T)\},
\end{align*}
{and in particular, we define $L_n(\eta) = \mathbb{P}_n\{\Delta l_n(T,\mathbf{v},\eta)\}$ and   $L_0(\eta) = \mathbb{P}\{ \Delta l_0(T,\mathbf{v},\eta)\}$.}
Here, the expectation is taken with respect to the joint distribution of $T, \Delta$ and $\mathbf{v}$.

\section{Proof of Theorem 1}
Define $\alpha_n = \gamma_n\log^2 n + a_n = \tau_n + a_n$. For some $D > 0$, let $\mathbb{R}_D^p\coloneqq\{ \boldsymbol{\beta} \in \mathbb{R}^p: \Vert \boldsymbol{\beta}\Vert_{\infty} < D\}$ and $\mathcal{G}_D\coloneqq \mathcal{G}(L,\mathbf{p},s,D)$, and define
\[
\hat{\eta}_D = \argmax_{\eta \in \mathbb{R}^p_D \times \mathcal{G}_D} PL(\eta).
\]
Further, denote by $\hat{\eta} =(\hat{\boldsymbol{\beta}}, \hat{g})$ a local maximizer of $PL(\eta)$ over $\mathbb{R}^p\times \mathcal{G}$, that is, by setting $D=\infty$ in  $\mathbb{R}^p_D$ and $\mathcal{G}_D$. As in \cite{zhong2022deep}, it can be shown that 
if $ \max (||\beta||, ||g||_{\infty}) \rightarrow \infty$, $PL(\eta) \rightarrow -\infty$; hence,
when $D$ is sufficiently large, $\hat{\eta} = \hat{\eta}_D$ almost surely.  
Therefore, in the following, we  show that $d(\hat{\eta}_D,\eta_0) = O_p(\alpha_n)$,   when $D$ is   sufficiently large. 

To do so, it suffices to show that for any $\epsilon > 0$, there exists a $C$ such that
\begin{equation} \label{epsilon}
    \operatorname{P} \Bigg\{ \sup_{\eta \in \mathcal{N}_c} PL(\eta) < PL(\eta_0)\Bigg\} \geq 1 - \epsilon,
\end{equation}
where $\mathcal{N}_{c} = \{\eta \in  \mathbb{R}^p_D \times \mathcal{G}_D: d(\eta,\eta_0) = C\alpha_n\}$. If it holds, it implies with probability at least $1-\epsilon$ that there exists a $C>0$ such that  a local maximum exists and is inside the ball $\mathcal{N}_{c}$. 
Hence, there exists a local maximizer such that $d(\hat{\eta}, \eta_0)= O_p(\alpha_n)$.

Without loss of generality, we  assume that $\eta$ satisfies $\mathbb{E}\{ {\eta}(\mathbf{v})\} =\mathbb{E}\{ {\eta_0}(\mathbf{v})\}$, implying 
$\mathbb{E}\{g(\mathbf{z})\}=0$; if not, we can always centralize it. To see this, consider any $\eta = (\beta,g)$ in the ball $B_C = \{ \eta \in  \mathbb{R}^p_D \times \mathcal{G}_D: d(\eta,\eta_0) \leq C\alpha_n\}$, its centralization $\eta' = (\beta,g - \mathbb{E}\{ {\eta}(\mathbf{v}) -  {\eta_0}(\mathbf{v})\})$ is also in the ball $B_C$,  satisfying $\mathbb{E}\{ {\eta'}(\mathbf{v})\} =\mathbb{E}\{ {\eta_0}(\mathbf{v})\}$ and $PL(\eta') = PL(\eta)$.    


{Because of the sparsity of the $\beta$-coefficients, we arrange the indices of the covariates $(x_1, \ldots, x_p)$ so that $\beta_{j0}=0$ when $j > s_\beta$.} We consider 
\begin{eqnarray}\label{D_eta}
    &  & PL(\eta) - PL(\eta_0) \nonumber \\
    & = & \{L_n(\eta) - L_n(\eta_0)\} - \sum_{j=1}^p\{p_{\lambda}(|\beta_{j}|) - p_{\lambda}(| \beta_{j0}|)\} \nonumber \\ 
    &\leq & \{L_n(\eta) - L_n(\eta_0)\} - \sum_{j=1}^{s_{\beta}}\{p_{\lambda}(|\beta_{j} |) - p_{\lambda}(| \beta_{j0}|)\}, 
    \end{eqnarray}
where the inequality holds because $p_{\lambda}(|\beta_j |) - p_{\lambda}(0) >0$
when $j > s_\beta$.

We first deal with
\begin{align}\label{partial likelihood ineq}
    \begin{split}
        L_n(\eta) - L_n(\eta_0) =& \{ L_0(\eta) - L_0(\eta_0)\}  \\
        & + \{L_n(\eta) - L_0(\eta )\}- \{ L_n(\eta_0) - L_0(\eta_0)\}.
    \end{split}
\end{align}
According to Lemma 2 in \cite{zhong2022deep}, we know that
\[
L_0(\eta) - L_0(\eta_0) \asymp -d^2(\eta, \eta_0).
\]
Since $d( \eta,\eta_0) = C\alpha_n$, the first term in the right hand side of \ref{partial likelihood ineq} is of the order $C^2 \alpha_n^2.$

After some calculation,  
\begin{align}
    \begin{split}
        (L_n-L_0)(\eta) -  (L_n-L_0)(\eta_0) =&(\mathbb{P}_n - \mathbb{P})\{\Delta l_0(T,\mathbf{v},\eta) -\Delta l_0(T,\mathbf{v},\eta_0) \}\\
        &+\mathbb{P}_n\Big\{\Delta\log\frac{S_0(T,\eta)}{S_0(T,\eta_0)} -   \Delta\log\frac{S_{0n}(T,\eta)}{S_{0n}(T,\eta_0)}\Big\}\\
        =&I + II.
    \end{split}
\end{align}

According to the proof of Theorem 3.1 in \cite{zhong2022deep}, with $\mathcal{A}_{\delta} = \{(\beta,g)\in \mathbb{R}^p_D \times \mathcal{G}_D:\delta/2 \leq d(\eta,\eta_0)\leq \delta \}$,  
it follows that
\begin{align*}
    \sup_{\eta \in \mathcal{A}_{\delta}}|I| &= O(n^{-1/2}\phi_n(\delta)), \\
    \sup_{\eta \in \mathcal{A}_{\delta}}|II | &\leq O(n^{-1/2}\phi_n(\delta)),
\end{align*}
where  $\phi_n(\delta) = \delta\sqrt{s\log\frac{{\cal U}}{\delta}} + \frac{s}{\sqrt{n}}\log\frac{{\cal U}}{\delta}$ and ${\cal U} = L\prod_{l=1}^L(p_l + 1)\sum_{l=1}^Lp_lp_{l+1}$.Then by Assumption 1, when $\delta = C(\tau_n + a_n)$, we can show that $n^{-1/2}\phi_n\{C(\tau_n + a_n)\} \leq C (\tau_n + a_n)^2 = C\alpha_n^2$.

By the Taylor expansion and the Cauchy-Schwarz inequality, the second term on
the right-hand side of (\ref{D_eta}) is bounded by
\[
\sqrt{s_{\beta}} a_n\Vert \boldsymbol{\beta} - \boldsymbol{\beta_0} \Vert + \frac{1}{2}b_n \Vert \boldsymbol{\beta} - \boldsymbol{\beta_0}\Vert^2.
\]
Since $d(\eta,\eta_0) = C\alpha_n$, and therefore $\Vert \boldsymbol{\beta} - \boldsymbol{\beta_0} \Vert$ is of the order $C\alpha_n$. Hence, this upper bound is dominated by the first term in (\ref{partial likelihood ineq}) as $b_n \rightarrow 0$ by the assumption.

Therefore,  for any $\epsilon>0$, there exist sufficiently large $C, D>0$ so that (\ref{epsilon}) holds, and hence $d(\hat{\eta}_D,\eta_0) = O_p(\alpha_n)$,  which gives $d(\hat{\eta},\eta_0) = O_p(\alpha_n)$, where we recall $\hat{\eta}$ is the local maximizer of $PL(\eta)$ over $\mathbb{R}^p\times \mathcal{G}$. We note that
 \begin{align*}
    d^2(\hat{\eta},\eta_0) &=\mathbb{E}[(\hat{\boldsymbol{\beta}}- \boldsymbol{\beta}_0)^{\top}\{\mathbf{x} - \mathbb{E}(\mathbf{x}|\mathbf{z})\} + (\hat{\boldsymbol{\beta}} - \boldsymbol{\beta}_0)^{\top}\mathbb{E}(\mathbf{x}|\mathbf{z}) + \{\hat{g}(\mathbf{z})-g_0(\mathbf{z}) \}]^2\\
    &=\mathbb{E}[(\hat{\boldsymbol{\beta}} - \boldsymbol{\beta}_0)^{\top}\{\mathbf{x} - \mathbb{E}(\mathbf{x}|\mathbf{z})\}]^2 + \mathbb{E}[\{\hat{g}(\mathbf{z})-g_0(\mathbf{z}) \} + (\hat{\boldsymbol{\beta}} - \boldsymbol{\beta}_0)^{\top}\mathbb{E}(\mathbf{x}|\mathbf{z})]^2,
\end{align*}
where the second equality holds because,  by the definition of  $d(\cdot, \cdot)$, 
$\mathbb{E}$ is taken with respect to the joint density of  $\mathbf{v}=(\mathbf{x}^\top,\mathbf{z}^\top)^\top$, which is independent of the observed data, and hence, $\hat{\boldsymbol{\beta}}$ and $\hat{g}$.  By Assumptions 2-4, it follows $\Vert \hat{\boldsymbol{\beta}} - \boldsymbol{\beta}_0 \Vert = O_p(\alpha_n)$ and $\Vert \hat{g} - g_0 \Vert_{L^2} = O_p(\alpha_n)$.

\section{Proof of Theorem 2}
For the claims made in Theorem 2, it suffices to show that, with probability tending to 1, for any given $\eta= (\boldsymbol{\beta},g)$ satisfying that $|| \eta -\eta_{0} || = O(\gamma_n\log^2 n)$, where 
$\eta_0= (\boldsymbol{\beta}_0,g_0)$, and some constant $C>0$, 
\[
PL\{(\boldsymbol{\beta}_1^{\top},\mathbf{0}^{\top})^{\top}, g \} = \max_{\Vert \boldsymbol{\beta}_2\Vert \leq C\gamma_n\log^2 n} PL\{(\boldsymbol{\beta}_1^{\top},\boldsymbol{\beta}_2^{\top})^{\top}, g  \},
\]
where $\boldsymbol{\beta}_1= (\beta_1, \ldots, \beta_{s_{\beta}})^{\top}$ and $\boldsymbol{\beta}_2= (\beta_{s_{\beta}+1}, \ldots, \beta_p)^{\top}$.
We only need to show that, for any $j = s_{\beta}+1,\dots,p$, 
\begin{align*}
    &\partial PL(\boldsymbol{\beta},g)/\partial\beta_j < 0, \quad \text{for } 0 < \beta_j < C\gamma_n\log^2 n;\\
    &\partial PL(\boldsymbol{\beta},g)/\partial\beta_j > 0, \quad \text{for } -C\gamma_n\log^2 n < \beta_j < 0.
\end{align*}

To proceed, we note that $\partial PL(\boldsymbol{\beta},g)/\partial\beta_j 
=  \partial \ell (\eta)/\partial\beta_j - sign(\beta_j)p'_{\lambda}(|\beta_j|)$
for $j = s_{\beta}+1,\dots,p$.
Denote by $F_j(\eta)$ the partial derivative of $\ell(\eta)$ w.r.t. $\beta_j$, i.e.
\begin{align*}
    F_j(\eta) &= \frac{\partial \ell(\eta)}{\partial \beta_j} = \frac{1}{n} \sum_{1 = 1}^n \int_0^{\tau} \Big\{ x_{i,j} -  \frac{\sum_{k=1}^n Y_k(s)x_{k,j}\exp(\boldsymbol{\beta}^{\top}\mathbf{x}_k + g(\mathbf{z}_k))}{\sum_{k=1}^n Y_k(s)\exp(\boldsymbol{\beta}^{\top}\mathbf{x}_k + g(\mathbf{z}_k))}\Big\}dN_i(s), 
\end{align*}
where $x_{k,j}$ (or  $x_{i,j}$) is the $j$th element of $\mathbf{x}_k$ (or $\mathbf{x}_i$).
As part of $\eta$ is a functional,  we consider a functional expansion of  $F_j(\eta)$ around  its truth, $\eta_0$.  Specifically, for a real number $0\le e \le 1$, we define ${\cal F}_j(e) = F_j\{\eta_0 + e(\eta - \eta_0) \}$, a function of the scalar $e$ only. Obviously,  ${\cal F}_j(1)=F_j(\eta)$ and ${\cal F}_j(0)=F_j(\eta_0)$.

Taking the Taylor expansion of ${\cal F}_j(1)$ around 0 gives
\begin{equation}  \label{deri}
{\cal F}_j(1) = {\cal F}_j(0) + {\cal F}_j'(0)+ {\cal F}_j''(e^*), 
\end{equation}
where $e^*$ is between 0 and 1. By some calculation, 
\begin{align*}
    &{\cal F}_j'(e) = -\frac{1}{n} \sum_{i =1}^n \int_0^{\tau} \Big[ \frac{\sum_k Y_k(s) \xi_e(\mathbf{v}_k)   x_{k,j}   (\eta-\eta_0)(\mathbf{v}_k)}{\sum_k Y_k(s) \xi_e(\mathbf{v}_k) } -\\
    & \frac{ \{\sum_k Y_k(s) \xi_e(\mathbf{v}_k)  x_{k,j} \}\{\sum_k Y_k(s)\xi_e(\mathbf{v}_k)  (\eta-\eta_0)(\mathbf{v}_k) \}}{\{\sum_k Y_k(s) \xi_e(\mathbf{v}_k) \}^2 } \Big] dN_i(s),
\end{align*}
where   $\mathbf{v}_k= (\mathbf{x}_k^{\top}, \mathbf{z}_k^{\top})^{\top}$, $\xi_e(\mathbf{v}_k) = \exp ( \{\eta_0 + e(\eta - \eta_0)\}(\mathbf{v}_k) ) $ and
$(\eta-\eta_0)(\mathbf{v}_k) =(\beta-\beta_0)^\top \mathbf{x}_k+ (g-g_0)(\mathbf{z}_k)$,
and  
\begin{align*}
    {\cal F}_j''(e) &= -\frac{1}{n} \sum_{i = 1}^n \int_0^{\tau} \Big[ \frac{\sum_k Y_k(s) \xi_e(\mathbf{v}_k) x_{k,j}  (\eta-\eta_0)^2(\mathbf{v}_k)}{\sum_k Y_k(s) \xi_e(\mathbf{v}_k)} \\
    &-   \frac{ 2\{\sum_k Y_k(s) \xi_e(\mathbf{v}_k)  x_{k,j}  (\eta-\eta_0)(\mathbf{v}_k) \}  \{\sum_k Y_k(s) \xi_e(\mathbf{v}_k)  (\eta-\eta_0)(\mathbf{v}_k) \}     }{ \{\sum_k Y_k(s)\xi_e(\mathbf{v}_k)\}^2} \\
    &-\frac{\{\sum_k Y_k(s)\xi_e(\mathbf{v}_k)x_{k,j} \} \{\sum_k Y_k(s)\xi_e(\mathbf{v}_k)(\eta - \eta_0)^2(\mathbf{v}_k) \}}{\{\sum_k Y_k(s) \xi_e(\mathbf{v}_k) \}^2} \\
    &+ \frac{2\{\sum_k Y_k(s)\xi_e(\mathbf{v}_k)x_{k,j} \}\{\sum_kY_k(s)\xi_e(\mathbf{v}_k)(\eta - \eta_0)(\mathbf{v}_k)\}^2}{\{\sum_k Y_k(s)\xi_e(\mathbf{v}_k) \}^3}
\Big] dN_i(s).
\end{align*}

It follows that ${\cal F}_j(0)$ in (\ref{deri}) is equal to
\begin{eqnarray*}
&  & \frac{1}{n} \sum_{i = 1}^n \int_0^{\tau} \Big\{ x_{i,j} -  \frac{\sum_{k=1}^n Y_k(s)x_{k,j}\exp(\boldsymbol{\beta}_0^{\top}\mathbf{x}_k + g_0(\mathbf{z}_k))}{\sum_{k=1}^n Y_k(s)\exp(\boldsymbol{\beta}_0^{\top}\mathbf{x}_k + g_0(\mathbf{z}_k))}\Big\}dN_i(s) \\
& = & \frac{1}{n} \sum_{1 = 1}^n \int_0^{\tau} \Big\{ x_{i,j} -  \frac{\sum_{k=1}^n Y_k(s)x_{k,j}\exp(\boldsymbol{\beta}_0^{\top}\mathbf{x}_k + g_0(\mathbf{z}_k))}{\sum_{k=1}^n Y_k(s)\exp(\boldsymbol{\beta}_0^{\top}\mathbf{x}_k + g_0(\mathbf{z}_k))}\Big\}dM_i(s),
\end{eqnarray*}
where $dM_i(s)= dN_i(s) - \lambda_0(s) Y_i(s) \exp(\boldsymbol{\beta}_0^{\top}\mathbf{x}_i + g_0(\mathbf{z}_i)) ds$ is the martingale with respect to the history up to time $s$.
Hence, $ n^{1/2}{\cal F}_j(0)$ converges in distribution to a normal distribution by the martingale central limit theorem \citep{fleming2013counting}, and therefore, ${\cal F}_j(0) = O_p(n^{-1/2})$.

We then consider 
\begin{eqnarray*}
    {\cal F}_j'(0) & = &  -\frac{1}{n} \sum_{i =1}^n \int_0^{\tau} \Big[ \frac{\sum_k Y_k(s) \xi_0(\mathbf{v}_k)   x_{k,j}   (\eta-\eta_0)(\mathbf{v}_k)}{\sum_k Y_k(s) \xi_0(\mathbf{v}_k) } \\
     & - & \frac{ \{\sum_k Y_k(s) \xi_0(\mathbf{v}_k)  x_{k,j} \}\{\sum_k Y_k(s)\xi_0(\mathbf{v}_k)  (\eta-\eta_0)(\mathbf{v}_k) \}}{\{\sum_k Y_k(s) \xi_0(\mathbf{v}_k) \}^2 } \Big] dN_i(s) \\
   & = &  -\frac{1}{n} \sum_{i =1}^n \int_0^{\tau} \Big[ \frac{\sum_k Y_k(s) \xi_0(\mathbf{v}_k)   x_{k,j}   (\eta-\eta_0)(\mathbf{v}_k)}{\sum_k Y_k(s) \xi_0(\mathbf{v}_k) } \\
    &- &  \frac{ \{\sum_k Y_k(s) \xi_0(\mathbf{v}_k)  x_{k,j} \}\{\sum_k Y_k(s)\xi_0(\mathbf{v}_k)  (\eta-\eta_0)(\mathbf{v}_k) \}}{\{\sum_k Y_k(s) \xi_0(\mathbf{v}_k) \}^2 } \Big] dM_i(s) \\
   &  -  &  \frac{1}{n} \sum_{i =1}^n \int_0^{\tau} \Big[ \frac{\sum_k Y_k(s) \xi_0(\mathbf{v}_k)   x_{k,j}   (\eta-\eta_0)(\mathbf{v}_k)}{\sum_k Y_k(s) \xi_0(\mathbf{v}_k) } \\
    & - & \frac{ \{\sum_k Y_k(s) \xi_0(\mathbf{v}_k)  x_{k,j} \}\{\sum_k Y_k(s)\xi_0(\mathbf{v}_k)  (\eta-\eta_0)(\mathbf{v}_k) \}}{\{\sum_k Y_k(s) \xi_0(\mathbf{v}_k) \}^2 } \Big] Y_i(s) \xi_0(\mathbf{v}_i) \lambda_0(s) ds \\
    & = & I_1 + I_2,
\end{eqnarray*}
where $\xi_0(\mathbf{v}_k)= \exp ( \eta_0(\mathbf{v}_k))=\exp(\boldsymbol{\beta}_0^{\top}\mathbf{x}_i + g_0(\mathbf{z}_i)).$
It follows that each summed item in  $I_1$,  i.e., 
\begin{eqnarray*}
&  & \int_0^{\tau} \Big[ \frac{\sum_k Y_k(s) \xi_0(\mathbf{v}_k)   x_{k,j}   (\eta-\eta_0)(\mathbf{v}_k)}{\sum_k Y_k(s) \xi_0(\mathbf{v}_k) }  \\
  &   -  &   \frac{ \{\sum_k Y_k(s) \xi_0(\mathbf{v}_k)  x_{k,j} \}\{\sum_k Y_k(s)\xi_0(\mathbf{v}_k)  (\eta-\eta_0)(\mathbf{v}_k) \}}{\{\sum_k Y_k(s) \xi_0(\mathbf{v}_k) \}^2 } \Big] dM_i(s),
  \end{eqnarray*}  
is a square integrable martingale \citep{fleming2013counting}. Hence, by the law of large numbers for martingales \citep{hall2014martingale}, $I_1 \rightarrow 0$ in probability.

Also, $I_2$ can be shown to be equal to
\begin{eqnarray*}
&  & -\int_0^{\tau} \frac{1}{n} \Big[ {\sum_k Y_k(s) \xi_0(\mathbf{v}_k)   x_{k,j}   (\eta-\eta_0)(\mathbf{v}_k)}  \\
  &  -  &  \frac{ \{\sum_k Y_k(s) \xi_0(\mathbf{v}_k)  x_{k,j} \}\{\sum_k Y_k(s)\xi_0(\mathbf{v}_k)  (\eta-\eta_0)(\mathbf{v}_k) \}}{ \sum_k Y_k(s) \xi_0(\mathbf{v}_k)  } \Big] \lambda_0(s)ds. 
\end{eqnarray*}
Define
\begin{eqnarray*}
S_{x_j,1}(s, \eta-\eta_0) =  \mathbb{E}[ Y_k(s) \xi_0(\mathbf{v}_k)   x_{k,j}   (\eta-\eta_0)(\mathbf{v}_k)], & S_{x_j}(s,\eta_0) =   \mathbb{E}[Y_k(s) \xi_0(\mathbf{v}_k)  x_{k,j}],\\
S_{1}(s, \eta-\eta_0)  =  \mathbb{E} [Y_k(s)\xi_0(\mathbf{v}_k)  (\eta-\eta_0)(\mathbf{v}_k)], & 
 S_{0}(s,\eta_0)  =   \mathbb{E} [Y_k(s)\xi_0(\mathbf{v}_k)].
\end{eqnarray*} 
Applying the empirical process arguments \citep{pollard1990empirical, wellner2005empirical}  yields that 
\begin{eqnarray*}
& & \sup_{s \in [0, \tau]} \big |\frac{1}{n} \Big[ {\sum_k Y_k(s) \xi_0(\mathbf{v}_k)   x_{k,j}   (\eta-\eta_0)(\mathbf{v}_k)}   \\
 &     -  &  \frac{ \{\sum_k Y_k(s) \xi_0(\mathbf{v}_k)  x_{k,j} \}\{\sum_k Y_k(s)\xi_0(\mathbf{v}_k)  (\eta-\eta_0)(\mathbf{v}_k) \}}{\sum_k Y_k(s) \xi_0(\mathbf{v}_k) } \Big] \\
& -  & S_{x_j,1}(s, \eta-\eta_0) +\frac{  S_{x_j}(s,\eta_0) S_{1}(s, \eta-\eta_0)}{  S_{0}(s,\eta_0)} \big | \rightarrow 0
\end{eqnarray*}
in probability, which implies that
$$I_2  \rightarrow - \int_0^\tau \left \{S_{x_j,1}(s, \eta-\eta_0) - \frac{  S_{x_j}(s,\eta_0) S_{1}(s, \eta-\eta_0)}{  S_{0}(s,\eta_0)}\right\} \lambda_0(s)ds$$ in probability.
Collecting all these terms, we thus have that
$${\cal F}_j'(0) =  - \int_0^\tau \left \{S_{x_j,1}(s, \eta-\eta_0) - \frac{  S_{x_j}(s,\eta_0) S_{1}(s, \eta-\eta_0)}{  S_{0}(s,\eta_0)}\right\} \lambda_0(s)ds + o_p(1).$$

We now bound  ${\cal F}_j'(0)$.  First note that, for any $s \in [0, \tau],$
\begin{eqnarray*}
     S_{x_j,1}(s, \eta-\eta_0) & \le &  \mathbb{E}[ \xi_0(\mathbf{v}_k)  | x_{k,j} |  | (\eta-\eta_0)(\mathbf{v}_k)|] \\
       &\le & \max_{ \mathbf{v}_k \in \mathbb{D}} \{\xi_0(\mathbf{v}_k)  | x_{k,j} | \} \int_\mathbb{D}  | (\eta-\eta_0)(\mathbf{v})| f_{\mathbf{v}_k}(\mathbf{v}) d\mathbf{v} \\
& \le  & C_1 \left \{\int_\mathbb{D}  (\eta-\eta_0)^2(\mathbf{v}) f_{\mathbf{v}_k}(\mathbf{v}) d\mathbf{v} \right\}^{1/2} \\
&= & C_1 ||\eta-\eta_0||,
\end{eqnarray*}
where $C_1>0 $  is a constant, $f_{\mathbf{v}_k}(\cdot)$ is the density function of the random vector $\mathbf{v}_k$, and  the last inequality stems from the Cauchy-Schwartz inequality,
in conjunction with  the boundedness assumptions on the covariates (i.e., $\mathbb{D}$ is bounded) and $\eta_0$ (Conditions 2 and 3 in the main text). Similarly,
we can show that, for any $s \in [0, \tau],$
\begin{eqnarray*}
    |S_{1}(s, \eta-\eta_0)|
\le C_2 ||\eta-\eta_0||, \, \,\,  |S_{x_j}(s,\eta_0)| \le C_3,  \, \,\,   S_{0}(s,\eta_0) \ge C_4,
\end{eqnarray*}
where $C_2, C_3, C_4>0$ are constants. The last inequality holds because at $\tau$, there is  at least probability of $\delta>0$ of observing subjects at risk (Condition 4 in the  main text), implying that $\min_{s \in [0, \tau]} \mathbb{E} (Y_k(s) | \mathbf{v}_k) \ge  \delta>0$ a.s.

As  \begin{eqnarray*}
&  &  | \int_0^\tau \left \{S_{x_j,1}(s, \eta-\eta_0) - \frac{  S_{x_j}(s,\eta_0) S_{1}(s, \eta-\eta_0)}{  S_{0}(s,\eta_0)}\right\} \lambda_0(s)ds|  \\
& \le & \int_0^\tau  |\{S_{x_j,1}(s, \eta-\eta_0)| \lambda_0(s)ds
+  \int_0^\tau \frac{  |S_{x_j}(s,\eta_0)| |S_{1}(s, \eta-\eta_0)|}{  S_{0}(s,\eta_0)} \lambda_0(s)ds \\
& \le & (C_1 + C_2 C_3 C_4^{-1}) \Lambda_0(\tau) ||\eta-\eta_0||, 
\end{eqnarray*}
where $\Lambda_0(\tau) = \int_0^\tau \lambda_0(s)ds < \infty$. Therefore, ${\cal F}_j'(0) = O_p( ||\eta-\eta_0||)$.

Similarly, using the explicit form of ${\cal F}_j''(e)$, some calculation  can show that 
${\cal F}_j''(e^*) = o_p(||\eta-\eta_0||)$.
Then we conclude that
\begin{eqnarray*}
\partial PL(\boldsymbol{\beta},g)/\partial\beta_j 
& =  &  \partial \ell (\eta)/\partial\beta_j - sign(\beta_j)p'_{\lambda}(|\beta_j|) \\
& =& \lambda [ \lambda^{-1}  ( {\cal F}_j(0) + {\cal F}_j'(0)+ {\cal F}_j''(e^*)) - sign(\beta_j) \lambda^{-1} p'_{\lambda}(|\beta_j|)].
\end{eqnarray*}

With the assumptions of $\lambda^{-1} \gamma_n \log^{2}(n) \rightarrow 0$
and $\lambda^{-1} n^{-1/2} \rightarrow 0$, 
it follows that  
$$\lambda^{-1}  ( {\cal F}_j(0) + {\cal F}_j'(0)+ {\cal F}_j''(e^*)) =o_p(1).$$ On the other hand, using the condition  
of $ \liminf_{n \rightarrow \infty}   \liminf_{\theta \rightarrow 0+}  \lambda^{-1} p'_{\lambda}(\theta)>0$, it follows that the sign of $\partial PL(\boldsymbol{\beta},g)/\partial\beta_j$ is the opposite sign of $\beta_j$ with probability going to 1. Hence, the claims follow.

\begin{figure}[H]
     \centering
    \subfloat[Selection of $\lambda$ for 10 simulated datasets]{\label{BIC}{\includegraphics[width=0.5\textwidth]{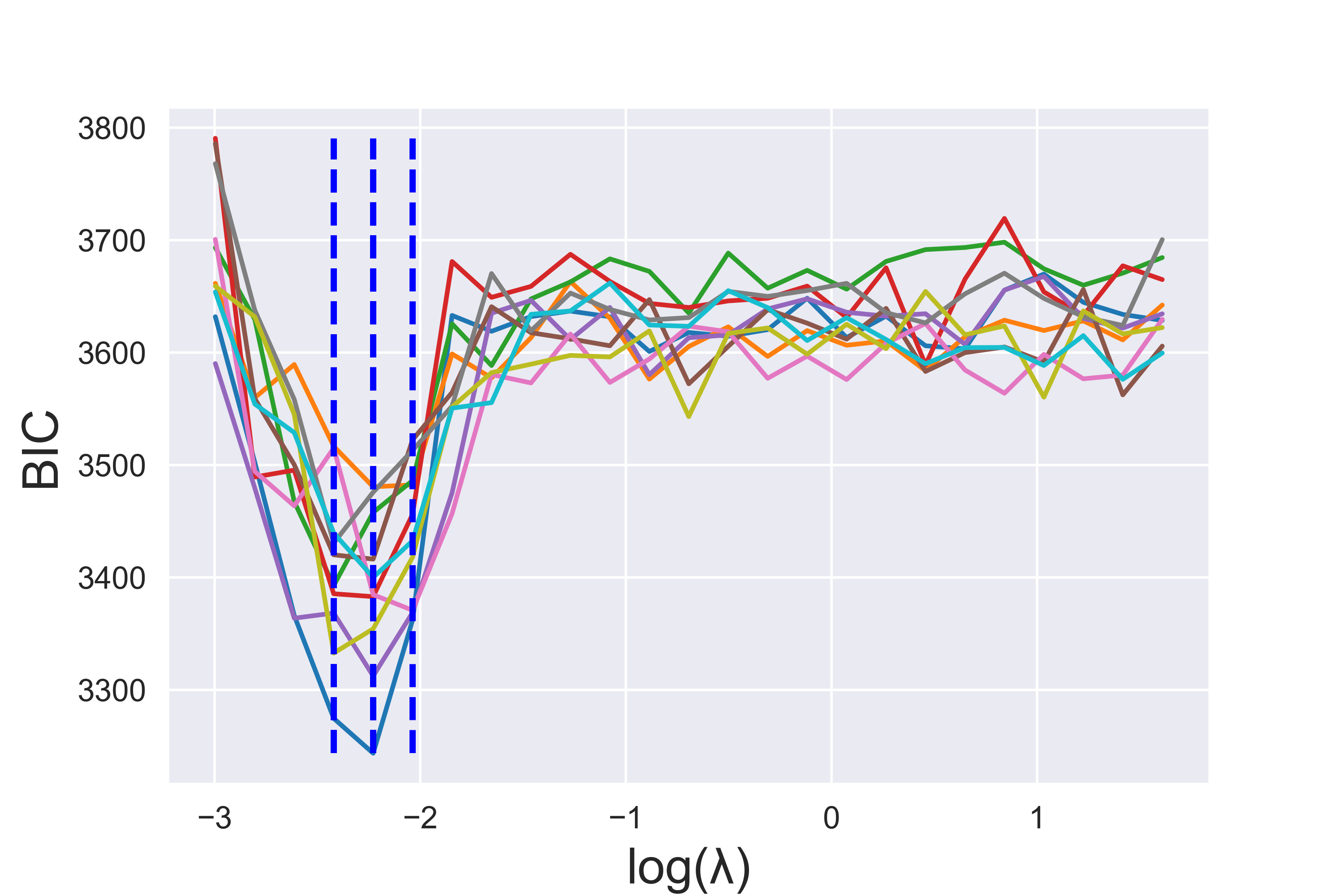}}}\hfill
    \subfloat[Selection path for the non-zero coefficients]{\label{coef_path}{\includegraphics[width=0.5\textwidth]{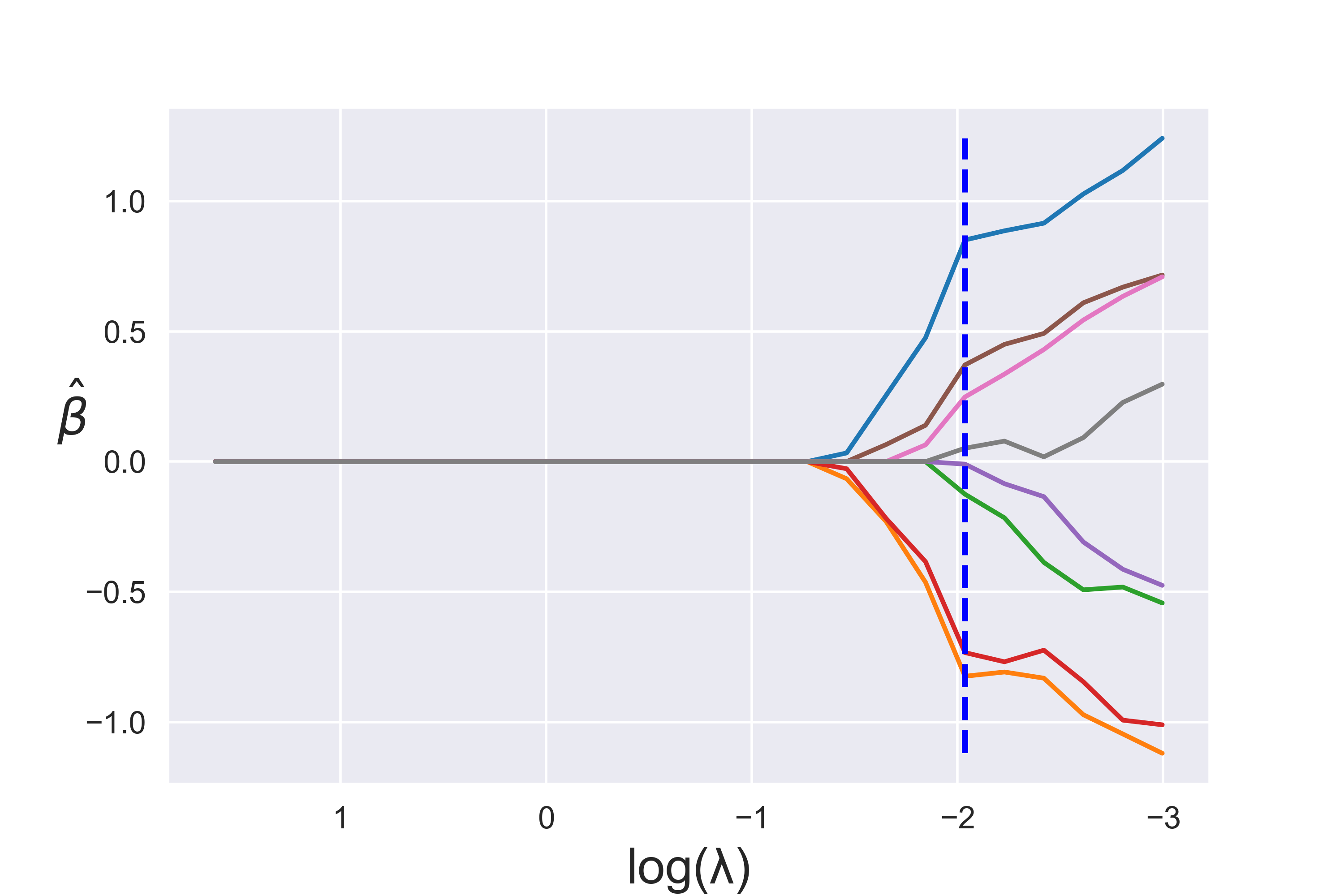}}}
     \caption{\textbf{Selection of} $\boldsymbol{\lambda}$ \textbf{in Penalized DPLC using BIC.} 
     }
\end{figure}

\begin{figure}
    \centering
    \includegraphics[scale = 0.85]{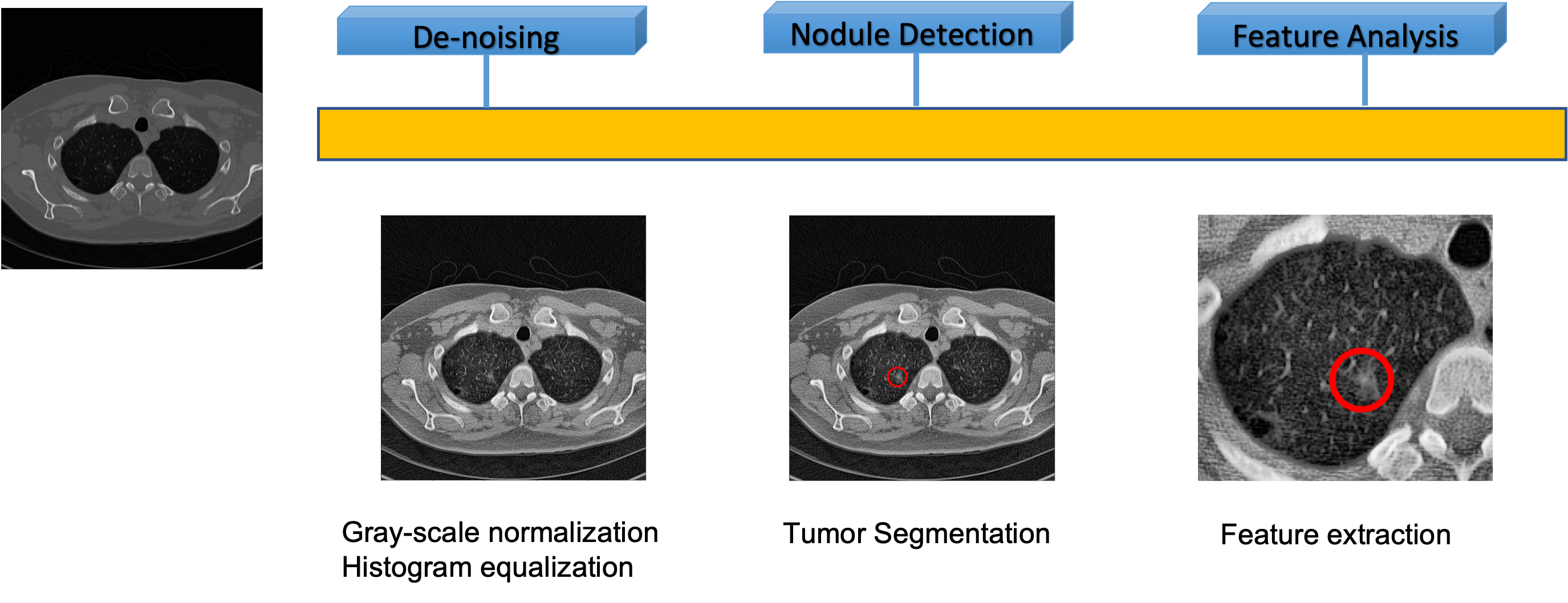}
    \caption{\textbf{Image Preprocessing Pipeline}}
    \label{image preprocess}
\end{figure}

\begin{figure}
    \centering
    \includegraphics[scale =0.45]{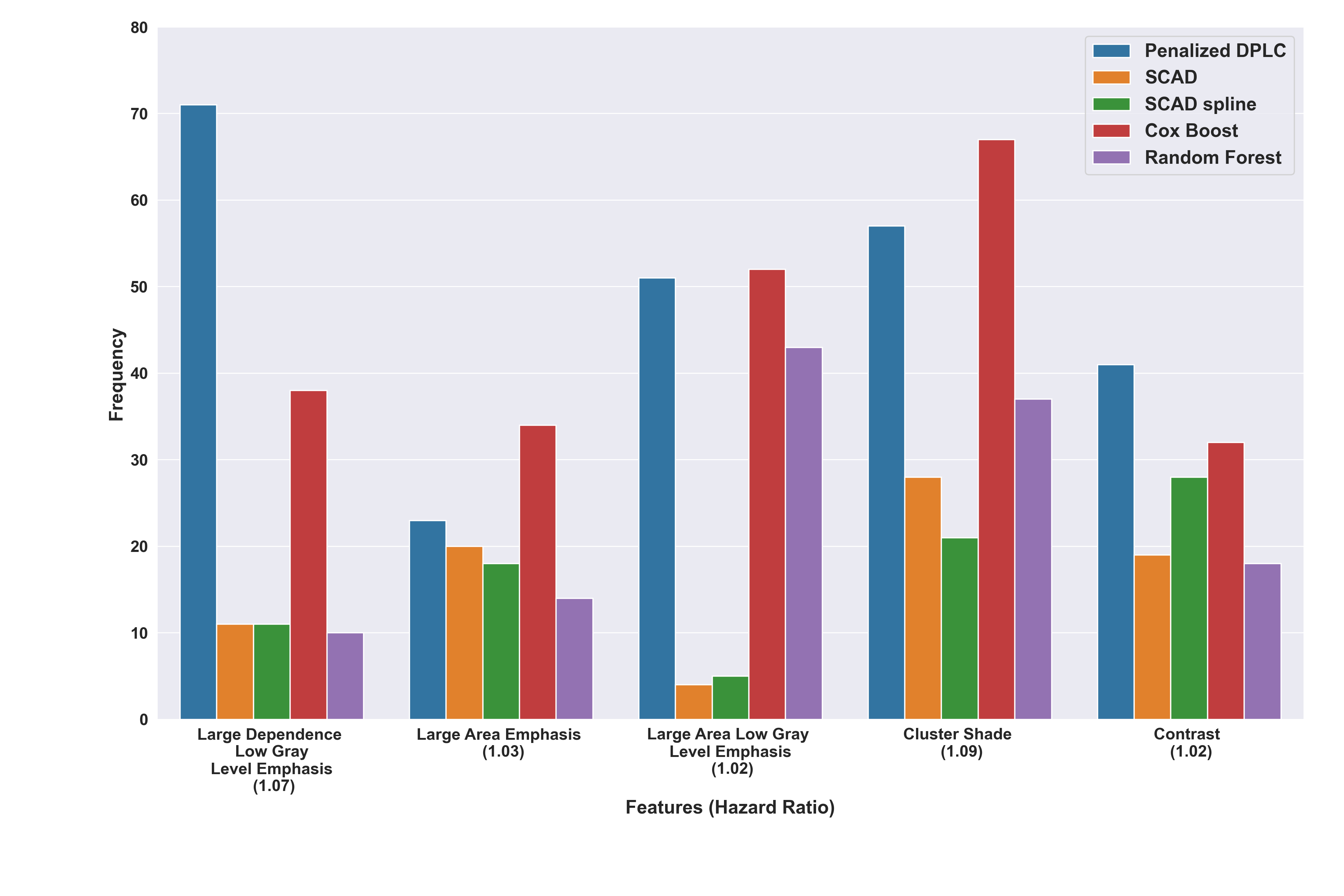}
    \caption{\textbf{Selection Frequency and Hazard Ratio of Selected Features:} The selection frequency of the most frequently selected five texture features is reported. The hazard ratio is the average of 100 experiments}
    \label{select_freq}
\end{figure}

\end{appendix}

\bibliographystyle{unsrtnat}
\bibliography{references}  

\begin{thebibliography}{36}
\providecommand{\natexlab}[1]{#1}
\providecommand{\url}[1]{\texttt{#1}}
\expandafter\ifx\csname urlstyle\endcsname\relax
  \providecommand{\doi}[1]{doi: #1}\else
  \providecommand{\doi}{doi: \begingroup \urlstyle{rm}\Url}\fi

\bibitem[Bade and Cruz(2020)]{bade2020lung}
Brett~C Bade and Charles S~Dela Cruz.
\newblock Lung cancer 2020: epidemiology, etiology, and prevention.
\newblock \emph{Clinics in Chest Medicine}, 41\penalty0 (1):\penalty0 1--24,
  2020.

\bibitem[Barbeau et~al.(2006)Barbeau, Li, Calderon, Hartman, Quinn, Markkanen,
  Roelofs, Frazier, and Levenstein]{barbeau2006results}
Elizabeth~M Barbeau, Yi~Li, Patricia Calderon, Cathy Hartman, Margaret Quinn,
  Pia Markkanen, Cora Roelofs, Lindsay Frazier, and Charles Levenstein.
\newblock Results of a union-based smoking cessation intervention for
  apprentice iron workers.
\newblock \emph{Cancer Causes \& Control}, 17:\penalty0 53--61, 2006.

\bibitem[Team(2011)]{national2011reduced}
National Lung Screening Trial~Research Team.
\newblock Reduced lung-cancer mortality with low-dose computed tomographic
  screening.
\newblock \emph{New England Journal of Medicine}, 365\penalty0 (5):\penalty0
  395--409, 2011.

\bibitem[Lubner et~al.(2017)Lubner, Smith, Sandrasegaran, Sahani, and
  Pickhardt]{lubner2017ct}
Meghan~G Lubner, Andrew~D Smith, Kumar Sandrasegaran, Dushyant~V Sahani, and
  Perry~J Pickhardt.
\newblock Ct texture analysis: definitions, applications, biologic correlates,
  and challenges.
\newblock \emph{Radiographics}, 37\penalty0 (5):\penalty0 1483--1503, 2017.

\bibitem[Lambin et~al.(2017)Lambin, Leijenaar, Deist, Peerlings, De~Jong,
  Van~Timmeren, Sanduleanu, Larue, Even, Jochems, et~al.]{lambin2017radiomics}
Philippe Lambin, Ralph~TH Leijenaar, Timo~M Deist, Jurgen Peerlings, Evelyn~EC
  De~Jong, Janita Van~Timmeren, Sebastian Sanduleanu, Ruben~THM Larue, Aniek~JG
  Even, Arthur Jochems, et~al.
\newblock Radiomics: the bridge between medical imaging and personalized
  medicine.
\newblock \emph{Nature Reviews Clinical Oncology}, 14\penalty0 (12):\penalty0
  749--762, 2017.

\bibitem[Cox(1972)]{Cox1972regression}
David~R Cox.
\newblock Regression models and life-tables.
\newblock \emph{Journal of the Royal Statistical Society: Series B
  (Methodological)}, 34\penalty0 (2):\penalty0 187--202, 1972.

\bibitem[Huang(1999)]{huang1999efficient}
Jian Huang.
\newblock Efficient estimation of the partly linear additive {Cox} model.
\newblock \emph{The Annals of Statistics}, 27\penalty0 (5):\penalty0
  1536--1563, 1999.

\bibitem[Zhong et~al.(2022)Zhong, Mueller, and Wang]{zhong2022deep}
Qixian Zhong, Jonas Mueller, and Jane-Ling Wang.
\newblock Deep learning for the partially linear {Cox} model.
\newblock \emph{The Annals of Statistics}, 50\penalty0 (3):\penalty0
  1348--1375, 2022.

\bibitem[Leshno et~al.(1993)Leshno, Lin, Pinkus, and
  Schocken]{leshno1993multilayer}
Moshe Leshno, Vladimir~Ya Lin, Allan Pinkus, and Shimon Schocken.
\newblock Multilayer feedforward networks with a nonpolynomial activation
  function can approximate any function.
\newblock \emph{Neural Networks}, 6\penalty0 (6):\penalty0 861--867, 1993.

\bibitem[Schmidt-Hieber(2020)]{schmidt2020nonparametric}
Johannes Schmidt-Hieber.
\newblock Nonparametric regression using deep neural networks with relu
  activation function.
\newblock \emph{The Annals of Statistics}, 48\penalty0 (4):\penalty0
  1875--1897, 2020.

\bibitem[Li et~al.(2020)Li, Soltanolkotabi, and Oymak]{li2020gradient}
Mingchen Li, Mahdi Soltanolkotabi, and Samet Oymak.
\newblock Gradient descent with early stopping is provably robust to label
  noise for overparameterized neural networks.
\newblock In \emph{International conference on artificial intelligence and
  statistics}, pages 4313--4324. PMLR, 2020.

\bibitem[Srivastava et~al.(2014)Srivastava, Hinton, Krizhevsky, Sutskever, and
  Salakhutdinov]{srivastava2014dropout}
Nitish Srivastava, Geoffrey Hinton, Alex Krizhevsky, Ilya Sutskever, and Ruslan
  Salakhutdinov.
\newblock Dropout: a simple way to prevent neural networks from overfitting.
\newblock \emph{The Journal of Machine Learning Research}, 15\penalty0
  (1):\penalty0 1929--1958, 2014.

\bibitem[Fan and Li(2001)]{fan2001variable}
Jianqing Fan and Runze Li.
\newblock Variable selection via nonconcave penalized likelihood and its oracle
  properties.
\newblock \emph{Journal of the American statistical Association}, 96\penalty0
  (456):\penalty0 1348--1360, 2001.

\bibitem[Fan and Li(2002)]{fan2002variable}
Jianqing Fan and Runze Li.
\newblock Variable selection for {Cox}'s proportional hazards model and frailty
  model.
\newblock \emph{The Annals of Statistics}, 30\penalty0 (1):\penalty0 74--99,
  2002.

\bibitem[Kingma and Ba(2014)]{kingma2014adam}
Diederik~P Kingma and Jimmy Ba.
\newblock Adam: A method for stochastic optimization.
\newblock \emph{arXiv preprint arXiv:1412.6980}, 2014.

\bibitem[Glorot and Bengio(2010)]{glorot2010understanding}
Xavier Glorot and Yoshua Bengio.
\newblock Understanding the difficulty of training deep feedforward neural
  networks.
\newblock In \emph{Proceedings of the thirteenth international conference on
  artificial intelligence and statistics}, pages 249--256. JMLR Workshop and
  Conference Proceedings, 2010.

\bibitem[Ji et~al.(2021)Ji, Li, and Telgarsky]{ji2021early}
Ziwei Ji, Justin Li, and Matus Telgarsky.
\newblock Early-stopped neural networks are consistent.
\newblock \emph{Advances in Neural Information Processing Systems},
  34:\penalty0 1805--1817, 2021.

\bibitem[Breheny and Huang(2011)]{breheny2011coordinate}
Patrick Breheny and Jian Huang.
\newblock Coordinate descent algorithms for nonconvex penalized regression,
  with applications to biological feature selection.
\newblock \emph{The Annals of Applied Statistics}, 5\penalty0 (1):\penalty0
  232, 2011.

\bibitem[Donoho and Johnstone(1994)]{donoho1994ideal}
David~L Donoho and Iain~M Johnstone.
\newblock Ideal spatial adaptation by wavelet shrinkage.
\newblock \emph{Biometrika}, 81\penalty0 (3):\penalty0 425--455, 1994.

\bibitem[Horowitz(2009)]{horowitz2009semiparametric}
Joel~L Horowitz.
\newblock \emph{Semiparametric and nonparametric methods in econometrics},
  volume~12.
\newblock Springer, 2009.

\bibitem[Hu and Lian(2013)]{hu2013variable}
Yuao Hu and Heng Lian.
\newblock Variable selection in a partially linear proportional hazards model
  with a diverging dimensionality.
\newblock \emph{Statistics \& Probability Letters}, 83\penalty0 (1):\penalty0
  61--69, 2013.

\bibitem[Binder et~al.(2009)Binder, Allignol, Schumacher, and
  Beyersmann]{binder2009boosting}
Harald Binder, Arthur Allignol, Martin Schumacher, and Jan Beyersmann.
\newblock Boosting for high-dimensional time-to-event data with competing
  risks.
\newblock \emph{Bioinformatics}, 25\penalty0 (7):\penalty0 890--896, 2009.

\bibitem[Ishwaran et~al.(2008)Ishwaran, Kogalur, Blackstone, and
  Lauer]{ishwaran2008random}
Hemant Ishwaran, Udaya~B Kogalur, Eugene~H Blackstone, and Michael~S Lauer.
\newblock Random survival forests.
\newblock \emph{The Annals of Applied Statistics}, 2\penalty0 (3):\penalty0
  841--860, 2008.

\bibitem[Katzman et~al.(2018)Katzman, Shaham, Cloninger, Bates, Jiang, and
  Kluger]{katzman2018deepsurv}
Jared~L Katzman, Uri Shaham, Alexander Cloninger, Jonathan Bates, Tingting
  Jiang, and Yuval Kluger.
\newblock Deepsurv: personalized treatment recommender system using a {Cox}
  proportional hazards deep neural network.
\newblock \emph{BMC Medical Research Methodology}, 18\penalty0 (1):\penalty0
  1--12, 2018.

\bibitem[Amadasun and King(1989)]{amadasun1989textural}
Moses Amadasun and Robert King.
\newblock Textural features corresponding to textural properties.
\newblock \emph{IEEE Transactions on systems, man, and Cybernetics},
  19\penalty0 (5):\penalty0 1264--1274, 1989.

\bibitem[Kong et~al.(2013)Kong, Akakin, and Sarma]{kong2013Generalized}
Hui Kong, Hatice~Cinar Akakin, and Sanjay~E Sarma.
\newblock A generalized laplacian of gaussian filter for blob detection and its
  applications.
\newblock \emph{IEEE transactions on cybernetics}, 43\penalty0 (6):\penalty0
  1719--1733, 2013.

\bibitem[Banerjee et~al.(2012)Banerjee, Moelker, Niessen, and
  Walsum]{banerjee20123d}
Jyotirmoy Banerjee, Adriaan Moelker, Wiro~J Niessen, and Theo~van Walsum.
\newblock 3d lbp-based rotationally invariant region description.
\newblock In \emph{Asian Conference on Computer Vision}, pages 26--37.
  Springer, 2012.

\bibitem[Tindle et~al.(2018)Tindle, Stevenson~Duncan, Greevy, Vasan, Kundu,
  Massion, and Freiberg]{tindle2018lifetime}
Hilary~A Tindle, Meredith Stevenson~Duncan, Robert~A Greevy, Ramachandran~S
  Vasan, Suman Kundu, Pierre~P Massion, and Matthew~S Freiberg.
\newblock Lifetime smoking history and risk of lung cancer: results from the
  framingham heart study.
\newblock \emph{JNCI: Journal of the National Cancer Institute}, 110\penalty0
  (11):\penalty0 1201--1207, 2018.

\bibitem[Lee and Giovannucci(2019)]{lee2019obesity}
Dong~Hoon Lee and Edward~L Giovannucci.
\newblock The obesity paradox in cancer: epidemiologic insights and
  perspectives.
\newblock \emph{Current Nutrition Reports}, 8:\penalty0 175--181, 2019.

\bibitem[Visbal et~al.(2004)Visbal, Williams, Nichols~III, Marks, Jett, Aubry,
  Edell, Wampfler, Molina, and Yang]{visbal2004gender}
Antonio~L Visbal, Brent~A Williams, Francis~C Nichols~III, Randolph~S Marks,
  James~R Jett, Marie-Christine Aubry, Eric~S Edell, Jason~A Wampfler, Julian~R
  Molina, and Ping Yang.
\newblock Gender differences in non--small-cell lung cancer survival: an
  analysis of 4,618 patients diagnosed between 1997 and 2002.
\newblock \emph{The Annals of Thoracic Surgery}, 78\penalty0 (1):\penalty0
  209--215, 2004.

\bibitem[Higgins et~al.(2012)Higgins, Chino, Ready, D’Amico, Berry, Sporn,
  Boyd, and Kelsey]{higgins2012lymphovascular}
Kristin~A Higgins, Junzo~P Chino, Neal Ready, Thomas~A D’Amico, Mark~F Berry,
  Thomas Sporn, Jessamy Boyd, and Chris~R Kelsey.
\newblock Lymphovascular invasion in non--small-cell lung cancer: implications
  for staging and adjuvant therapy.
\newblock \emph{Journal of Thoracic Oncology}, 7\penalty0 (7):\penalty0
  1141--1147, 2012.

\bibitem[Tibshirani(2011)]{tibshirani2011regression}
Robert Tibshirani.
\newblock Regression shrinkage and selection via the lasso: a retrospective.
\newblock \emph{Journal of the Royal Statistical Society: Series B (Statistical
  Methodology)}, 73\penalty0 (3):\penalty0 273--282, 2011.

\bibitem[Fleming and Harrington(2013)]{fleming2013counting}
Thomas~R Fleming and David~P Harrington.
\newblock \emph{Counting processes and survival analysis}, volume 625.
\newblock John Wiley \& Sons, 2013.

\bibitem[Hall and Heyde(2014)]{hall2014martingale}
Peter Hall and Christopher~C Heyde.
\newblock \emph{Martingale limit theory and its application}.
\newblock Academic press, 2014.

\bibitem[Pollard(1990)]{pollard1990empirical}
David Pollard.
\newblock Empirical processes: theory and applications.
\newblock Ims, 1990.

\bibitem[Wellner(2005)]{wellner2005empirical}
Jon~A Wellner.
\newblock Empirical processes: Theory and applications.
\newblock \emph{Notes for a course given at Delft University of Technology},
  page~17, 2005.

\end{thebibliography}






\end{document}